%% file: sample-acmlarge.tex
\RequirePackage[hyphens]{url}
\documentclass[acmlarge,nonacm]{acmart}
\usepackage{booktabs}
\usepackage{multirow}
\usepackage{adjustbox}
\usepackage{tabularx}
\newcolumntype{Y}{>{\centering\arraybackslash}X}

\usepackage{bbm}
\usepackage{graphicx, subcaption}

\usepackage{ctable}
\usepackage{array}
\newcolumntype{?}{!{\vrule width 1pt}}
\usepackage{bbding}
\usepackage{wasysym}
\usepackage{pifont}
\newcommand{\cmark}{\ding{51}}
\newcommand{\xmark}{\ding{55}}

\usepackage{commath}
\usepackage{booktabs}   
\newcommand{\ra}[1]{\renewcommand{\arraystretch}{#1}}
\usepackage{mathtools}

\usepackage{enumitem}
\usepackage[ruled]{algorithm2e}

\DontPrintSemicolon

\AtBeginDocument{%
  \providecommand\BibTeX{{%
    \normalfont B\kern-0.5em{\scshape i\kern-0.25em b}\kern-0.8em\TeX}}}

\setcopyright{acmcopyright}
\acmJournal{IMWUT}
\acmYear{2021} \acmVolume{5} \acmNumber{1} \acmArticle{2} \acmMonth{3} \acmPrice{15.00}\acmDOI{10.1145/3448090}

\begin{document}

\title{Lifelong Adaptive Machine Learning for Sensor-based Human Activity Recognition Using Prototypical Networks}

\author{Rebecca Adaimi}
\affiliation{%
  \institution{University of Texas at Austin}
  \streetaddress{2501 Speedway}
  \city{Austin}
  \state{Texas}
  \postcode{78712}
  \country{USA}}
  \email{rebecca.adaimi@utexas.edu}
    
\author{Edison Thomaz}
\affiliation{%
  \institution{University of Texas at Austin}
  \streetaddress{2501 Speedway}
  \city{Austin}
  \state{Texas}
  \postcode{78712}
  \country{USA}}  
  \email{ethomaz@utexas.edu}

\renewcommand{\shortauthors}{Adaimi et al.}

\begin{abstract}

Continual learning, also known as lifelong learning, is an emerging research topic that has been attracting increasing interest in the field of machine learning. With human activity recognition (HAR) playing a key role in enabling numerous real-world applications, an essential step towards the long-term deployment of such recognition systems is to extend the activity model to dynamically adapt to changes in people's everyday behavior. Current research in continual learning applied to HAR domain is still under-explored with researchers exploring existing methods developed for computer vision in HAR. Moreover, analysis has so far focused on task-incremental or class-incremental learning paradigms where task boundaries are known. This impedes the applicability of such methods for real-world systems since data is presented in a randomly streaming fashion. To push this field forward, we build on recent advances in the area of continual machine learning and design a lifelong adaptive learning framework using Prototypical Networks, \textit{LAPNet-HAR}, that processes sensor-based data streams in a task-free data-incremental fashion and mitigates catastrophic forgetting using experience replay and continual prototype adaptation. Online learning is further facilitated using contrastive loss to enforce inter-class separation. \textit{LAPNet-HAR} is evaluated on 5 publicly available activity datasets in terms of the framework's ability to acquire new information while preserving previous knowledge. Our extensive empirical results demonstrate the effectiveness of \textit{LAPNet-HAR} in task-free continual learning and uncover useful insights for future challenges.  

\end{abstract}



\maketitle
\input{main}

\bibliographystyle{ACM-Reference-Format}
\bibliography{bibliography}

\end{document}

%% file: main.tex
\section{Introduction}

Advances in mobile devices, sensors and computational methods over the last decade have brought Human Activity Recognition (HAR) to the forefront. Recently, HAR has begun to play a central role in enabling numerous emerging real-world applications ranging from mobile health \cite{10.1145/3301275.3302315,san2020eating} and disorder diagnosis \cite{cakmak2020using} to personal assistance and smart environments \cite{article, adaimi2021ok}. For instance, wearable sensors have been featured in automated dietary monitoring to track food intake gestures \cite{thomaz2015practical} and in health assessment applications to track sedentary activities of post-surgical patients \cite{king2015objective}. However, despite the progress the field has experienced, HAR is hampered by important learning shortcomings. HAR models today are typically built and evaluated with data offline under the assumption that the environment, people and their corresponding behaviors and activities remain constant over time. While this has been essential in establishing a methodological foundation for HAR and steady progress is made in the field, state-of-the-art approaches remain unable to close the gap between validation in controlled conditions and the realities and challenges of the real world. Life is dynamic, forcing people to change careers, move to a new city and develop new hobbies. Starting a family and adopting a pet introduces new habits and routines, and unexpected health issues might pose mobility and cognitive challenges that cause significant changes in how one performs everyday activities. A smart kitchen assistant should be able to recognize cooking activities even when individuals move to a new home. Similarly, a health monitoring system should be able to accurately track older adults’s behaviors and assess their ability to live independently even after the onset of impairments that cause them to move more slowly. Consider a smartphone application designed to monitor and provide just-in-time interventions to individuals suffering from post-traumatic stress disorder (PTSD). Ideally, the application should be adaptive and responsive to patients' behaviors, health states and medication changes over time. The COVID-19 outbreak provides another example of the need for human-centered systems that adapt to changing conditions. In a matter of months, the pandemic deeply altered human behaviors and routines around the world; millions of people changed jobs and social patterns, found new ways to exercise, learned new hobbies, and much more.

A key challenge in building adaptive HAR systems is that trained models must incorporate new knowledge while avoiding forgetting previously learned knowledge. Neural networks tend to exhibit loss of previously acquired knowledge when incorporating new information, a phenomenon known as \textit{catastrophic forgetting} (CF). Finding this balance between neural stability in remembering old classes and plasticity in learning new classes is the main hurdle for continual learning, which is a key ingredient for a long-term sustainable HAR system. To address this challenge, we leverage \textit{Continual Learning (CL)}, a learning paradigm that incrementally extending a neural network's acquired knowledge, whether learning new input domains or new classes and tasks. Continual learning, which has also been referred to as lifelong learning \cite{chen2018lifelong} or incremental learning \cite{polikar2001learn++}, has been attracting increasing interest in the field of machine learning, with researchers developing different methods for mitigating catastrophic forgetting when incrementally learning new tasks over time \cite{kemker2018measuring,parisi2019continual}. These approaches have achieved promising results in computer vision \cite{rebuffi2017icarl, delange2021continual} and robotics \cite{lesort2020continual}. 

While CL has shown much promise in the context of HAR \cite{a5929b3591af4a9f9338fcfc66e90c44, JHA2021}, emphasis has been given to CL approaches that are either task- or class-incremental. Task-incremental setting refers to data arriving sequentially in batches with one batch corresponding to one task, which is usually comprised of 2 or more classes. In other words, the model observes all data belonging to one task and is optimized for it before moving to another task. These assumptions confine the scope and applicability of a continual learning system since, in a real-world setting, it not only encounters new classes but also revisits old classes in a non-uniform dynamic way. Moreover, new activity classes are often introduced with a limited number of data samples, and the system needs to incrementally learn these new activities from a few samples in order to recognize them again at later times. With that in mind, we advocate for a framework that processes data in an \textit{online task-free incremental} fashion with the ability to incorporate new classes with limited data. Leveraging incremental prototypical networks \cite{50004}, we utilize metric-based learning to optimize class prototype representations in an online fashion. Prototype networks are compelling because they support new class representations with limited data as well as online prediction using nearest mean classification. Moreover, it provides a more general learning setting that eliminates the notion of tasks.

In this paper, we propose a \textbf{L}ifelong \textbf{A}daptive machine learning framework for HAR using \textbf{P}rototypical \textbf{Net}works, \textit{LAPNet-HAR}. We demonstrate that the framework (1) processes data streams in a task-free data-incremental online fashion, (2) incorporates a prototype memory scheme that evolves continually with the data stream enabling both online learning and prediction paradigm, (3) mitigates CF by employing experience replay using replay memory, and (4) prevents prototype obsolescence with replay-based prototype adaptation. We perform an extensive analysis of our framework across multiple datasets measuring not only performance but also trade-offs between learning new knowledge versus retaining previously acquired information. The contributions of this paper can be summarized as follows:

\begin{itemize}[noitemsep,topsep=3pt]
    \item A continual learning model for HAR using Prototypical Networks, \textit{LAPNet-HAR}, that incrementally evolves with streaming data in a task-free data-incremental paradigm and mitigates catastrophic forgetting using experience replay with continual prototype evolution. 
    \item An extensive evaluation of \textit{LAPNet-HAR} on 5 commonly used HAR datasets. The evaluation has demonstrated the effectiveness of \textit{LAPNet-HAR} in continual activity recognition as well as in handling the catastrophic forgetting and model intransigence tradeoff across all datasets. This further uncovered findings in dealing with various challenges of task-free continual learning in HAR.
\end{itemize}

\section{Related Work}
\label{sec:rel_work}
In this section, we review prior work in incremental/continual learning with a focus on sensor-based human activity recognition, as well as metric-based learning using prototypical networks.

\begin{table*}[t]
    \caption{Comparison of past CL paradigms applied to the HAR domain vs. our approach.}

     \centering
    \begin{tabular}{@{}lcccc@{}}

    \toprule 
        \textbf{Method} & \textbf{Continual Learning} & \textbf{Online Evaluation} & \textbf{Experience Replay}  & \textbf{Task-free Learning} \\

    \midrule
        CL-HAR-1 \cite{a5929b3591af4a9f9338fcfc66e90c44} & \cmark & \xmark & \xmark & \xmark \\
        CL-HAR-2 \cite{JHA2021} & \cmark & \xmark & \cmark & \xmark \\
        EMILEA \cite{EMILEA} & \cmark & \xmark & \cmark &  \xmark  \\
        HAR-GAN \cite{10.1145/3440036} & \cmark & \xmark & \cmark & \xmark \\
        \midrule
        LAPNet-HAR (ours) & \cmark & \cmark & \cmark & \cmark \\
        \bottomrule
    \end{tabular}
    \label{tab:rel_work_compare}
\end{table*}

\subsection{Continual Learning in HAR}

While most HAR-related prior work have mainly focused on offline supervised training, several methods with incremental learning abilities have been developed in the past years \cite{wang2012incremental,10.1145/3432230}. This incremental learning paradigm has not only been used to improve activity recognition performance \cite{ntalampiras2016incremental}, but also to personalize HAR models across users by adapting to distribution shifts \cite{siirtola2021context,10.1145/3432230,7850188}. For example, Learn++ was first introduced by Poliker \textit{et al.} \cite{polikar2001learn++} and used in \cite{mo2016human} to improve activity recognition systems when dealing with differences across individuals. Learn++ is an ensemble-based incremental learning algorithm that applies data weighting based on classification performance. There is also an increasing amount of work devoted to discovering and recognizing new activities in HAR. One popular approach is the use of clustering and one-class classifiers wherein a new cluster and classifier is added for every activity \cite{10.1145/3123021.3123044, 10.1145/3313831.3376875} enabling incremental learning. Gjoreski \textit{et al.} \cite{10.1145/3123021.3123044} applied an agglomerative clustering technique to perform real-time clustering of streaming data. Using temporal assumptions on human activities, short outliers were filtered out allowing a more accurate discovery of meaningful clusters. Wu \textit{et al.} \cite{10.1145/3313831.3376875} further extended this method with a self-supervised learning of activities labelled through one-shot interaction which optimizes the accuracy and user burden tradeoff.

However, simply creating a cluster or one-class classifier for every new activity while keeping data representation fixed can lead to the overlapping of clusters, which often would require re-training the model. Thus, another main component towards continual learning is evolving the HAR models with new data and activities. Fang \textit{et al.} \cite{8667728} employed a von Mises-Fisher-based hierarchical mixture model to model every type or pattern of activity as a component of the mixture model. Cheng \textit{et al.} \cite{10.1145/2462456.2464438} presented a zero-shot learner to adapt to new activities with limited training data. In this setup, semantic relationship between high-level activities and low-level sensor attributes are encoded using a knowledge-driven model; domain experts and developers must manually add new attributes and update the activity-attribute matrix for every new activity. 

All methods discussed thus far require a certain amount of re-engineering effort and therefore do not enable the automatic evolution of an activity model. For an end-to-end continual HAR system, a form of feedback loop is required where a new activity is discovered and fed into a network for evolution in an automatic and continual fashion while mitigating catastrophic forgetting of previously learned activities. Jha \textit{et al.} \cite{a5929b3591af4a9f9338fcfc66e90c44, JHA2021} explored whether continual learning techniques previously used in computer vision, and which possess these attributes, can be applied to HAR. They performed a comprehensive empirical analysis of several regularization-based and rehearsal-based approaches on various HAR datasets showing the challenges faced when moving towards sensor-based time series data. Moving towards dynamic architectures for activity models, Ye \textit{at al.} \cite{EMILEA} proposed EMILEA, a technique to evolve an activity model over time by integrating dynamic network expansion Net2Net \cite{45968} for enhancing the model's capacity with increasing number of activities and Gradient Episodic Memory \cite{NIPS2017_f8752278} for mitigating the effect of catastrophic forgetting. They further proposed another continual activity recognition system, HAR-GAN, that leverages Generative Adversarial Networks (GAN) to generate sensor data on previous classes without the need to store historical data \cite{10.1145/3440036}.

It is clear that the field is still under-explored and researchers studying continual learning in HAR have barely scratched the surface. To date, proposed methods, such as EMILEA \cite{EMILEA} and HAR-GAN \cite{10.1145/3440036}, have relied on dynamically expanding network architectures with increasing number of classes. While shown effective, this leads to increasing number of model parameters with the growing set of activities, which can be limiting for a deployable resource-constraint solution. In contrast with the typical softmax classifier, \textit{LAPNet-HAR} has a prediction head with fixed capacity, by applying representation learning of class prototypes for nearest-mean classification. This inherently supports a large number of classes without changing the network architecture. Moreover, while task-incremental learning has dominated continual learning HAR research, it is not a good match for real-world situations in which old activities are revisited and new activities are introduced in a non-uniform way and with limited training data. Thus, HAR systems with (1) continual, (2) online, and (3) task-free learning paradigm offer increased real-world practicality and applicability. A step forward towards such a system was presented in iCARL \cite{rebuffi2017icarl} for class-incremental learning. While it is also prototype-based, it requires task boundaries for knowledge distillation. Furthermore, iCARL applies a computationally expensive herding technique, which requires recalculation of the feature means on each change of the memory size or network parameters. \textit{LAPNet-HAR}, on the other hand, applies continual prototype adaptation which enables online prediction and also a contrastive representation loss that improves inter-class separation with the addition of new classes.

\subsection{Metric-based Learning Using Prototypical Networks}

Metric-based approaches aim to learn a set of embedding functions and a metric space that measures the similarity between support and target samples which in turn is used for classification. Researchers have proposed several ways to learn this metric space. For instance, Koch \textit{et al.} \cite{koch2015siamese} proposed a deep convolutional siamese network to compute a similarity score between the image pairs. The learned distance was then used to solve the one-shot learning problem via a K-nearest neighbors classification. Vinyals \textit{et al.} \cite{10.5555/3157382.3157504} presented an end-to-end trainable matching network that applies k-nearest neighbors using the cosine distance on the learned embedding features. Snell \textit{et al.} \cite{snell2017prototypical} extended the matching network and proposed the prototypical network that uses the Euclidean distance to learn prototype representations of each class. Breaking away from traditional few-shot learning, Ren \textit{et al.} \cite{ren18fewshotssl} extended the prototypical networks to semi-supervised few-shot learning. Incremental few-shot learning introduces the ability to jointly predict based on a set of previously learnt classes as well as additional novel classes \cite{gidaris2018dynamic,NEURIPS2019_e833e042}. Moving towards online contextualized few-shot learning, Ren \textit{et al.} \cite{50004} proposed using a contextual prototypical memory for encoding contextual information and storing previously learned classes. De Lange \textit{et al.} \cite{de2020continual} extended prototypical networks to an online data incremental learning setting with prototypes continually evolving with non-iid data streams.

Leveraging prototypical networks from online continual few-shot learning \cite{50004,de2020continual}, our work aims to build and analyze such a system for lifelong adaptive learning in HAR. Utilizing prototype memory with prototype evolution, our \textit{LAPNet-HAR} framework continually learns class representations and optimizes them using online averaging to account for distribution shift. Moreover, prototype adaptation is applied to accommodate changes in the latent space when new classes are introduced. This, in turn, prevents prototypes from becoming outdated (prototype obsolescence) which leads to incorrect class representation in the embedding space. \textit{To the best of our knowledge, we are also the first to analyze a continual learning system for HAR in a completely task-free continual setting}. Table \ref{tab:rel_work_compare} compares our proposed approach with past HAR-related CL work based on various properties of a continual learning paradigm.

\begin{figure}[t]
    \centering
    \includegraphics[width=0.9\columnwidth]{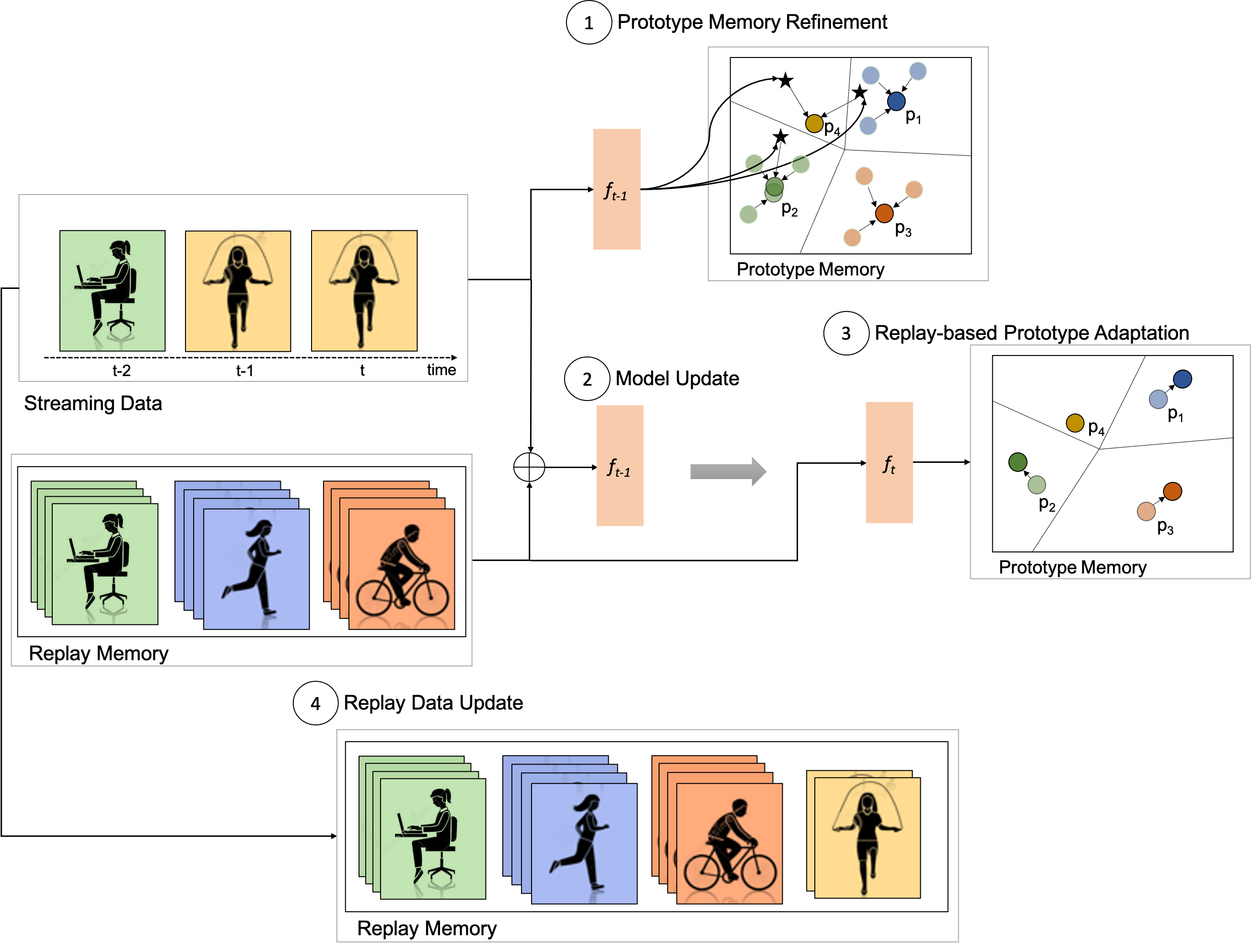}
    \caption{\textit{LAPNet-HAR} Overview. Given a pretrained model at time $t-1$ with a prototype memory containing prototypes of old classes as well as a replay buffer with a subset of old data samples, the new batch of streaming data is first used to update the prototype memory by using online averaging on old classes and creating prototypes of new classes (step 1). Then, with the updated prototypes, the data is combined with the replay data to update the model parameters (step 2). To prevent prototype obsolescence and adapt to the new latent space, the prototypes are adapted using the new embeddings of the replay data (step 3). Lastly, the replay memory is updated by random sampling from the current batch and extended to include new classes (step 4).}
    \label{fig:LAPNet-HAR}
\end{figure}{}

\begin{algorithm}[t]
\SetAlgoNoLine
\caption{LAPNet-HAR}
\KwIn{Initial network parameters $\theta_{0}$, hyperparameter $\alpha$, replay buffer size/class $b$}
\KwData{Training data $(X,Y)\ =\ D$}
$(X_0,Y_0)\ =\ D_0$ \tcp*{get base data for pretraining}
$C_{base} =$ set of base classes in $D_0$\;
$M_p\ =\ \{\};\ M_r\ =\ \{\}$ \tcp*{initialize empty prototype memory and replay buffer}
\;
\tcc{offline pretraining process}
Update $\theta_0$ with $(X_0,Y_0)$\;
Store prototypes $p_k$ for $k \in C_{base}$ in $M_p$ using $D_0$ \;
Sample data from $D_0$ and store in $M_r$ for replay\;
\;
\While {continually learning}{
$(X_t, Y_t)\ =\ D_t$\;
$M_p \leftarrow$ UpdatePrototypeMemory($D_t$, $M_p$, $\theta_{t-1}$) \tcp*{Figure \ref{fig:ProtoNEt_online} \& Equation \ref{eq:online_averaging}}
$Q \leftarrow D_t \cup M_r$; $(X_q,Y_q) = Q$  \tcp*{form combined query set}
Incur loss $\mathcal{L}(f_{\theta_{t-1}}(X_q),Y_q)$ \tcp*{Equation \ref{eq:loss_function}}
Update model $\theta_t$ with $(X_q,Y_q)$\;
$M_p \leftarrow$ PrototypeAdaptation($M_r$, $M_p$, $\theta_t$,$\alpha$) \tcp*{Equation \ref{eq:weighted_update}}
$M_r \leftarrow$ UpdateReplayBuffer($Q$, $M_r$, $\theta_t$, $b$)\;
}
\label{alg:LAPNet-HAR}
\end{algorithm}

\section{LAPNet-HAR Framework}

In this work, we introduce \textit{LAPNet-HAR} that enables continual adaptive learning for sensor-based data. We define continual learning in activity recognition as incrementally learning new patterns belonging to both old and novel activities in a sequential manner starting from a base model trained to classify a set of base classes. Figure \ref{fig:LAPNet-HAR} depicts an overview of \textit{LAPNet-HAR}. To reiterate, continual learning is enabled using (i) a prototypical network with prototype memory for incremental learning, (ii) experience replay of a small number of old classes’ data stored in a replay buffer to prevent catastrophic forgetting, (iii) continual replay-based prototype adaptation to prevent prototype obsolescence, and (iv) a contrastive loss \cite{hadsell2006dimensionality} that pushes samples from different classes apart to improve inter-class separation.

The workflow starts with an initial model $f_{\theta_0}$ and a set of base data $D_0 = (\{(x_i,y_i)\}_{i=1}^{N},C_{base})$ where, $x_i$ is a raw data frame with $y_i$ its corresponding label and $C_{base}$ is a set of base activities. During the continual learning process, a batch of training data $D_t = (X_t,Y_t)$ is presented at every time step $t$. The goal for \textit{LAPNet-HAR} is to continually learn and adapt a time-series sensor-based model $f_{\theta_t}$ with new incoming data while mitigating catastrophic forgetting. A secondary but important goal is learn with as few examples as possible. Algorithm \ref{alg:LAPNet-HAR} presents the overall \textit{LAPNet-HAR} framework. In the following, we give the algorithmic details of the key components of \textit{LAPNet-HAR}. 

\begin{figure}[t]
    \centering
    \begin{subfigure}[b]{0.49\linewidth}
    \centering
        \includegraphics[width=\columnwidth]{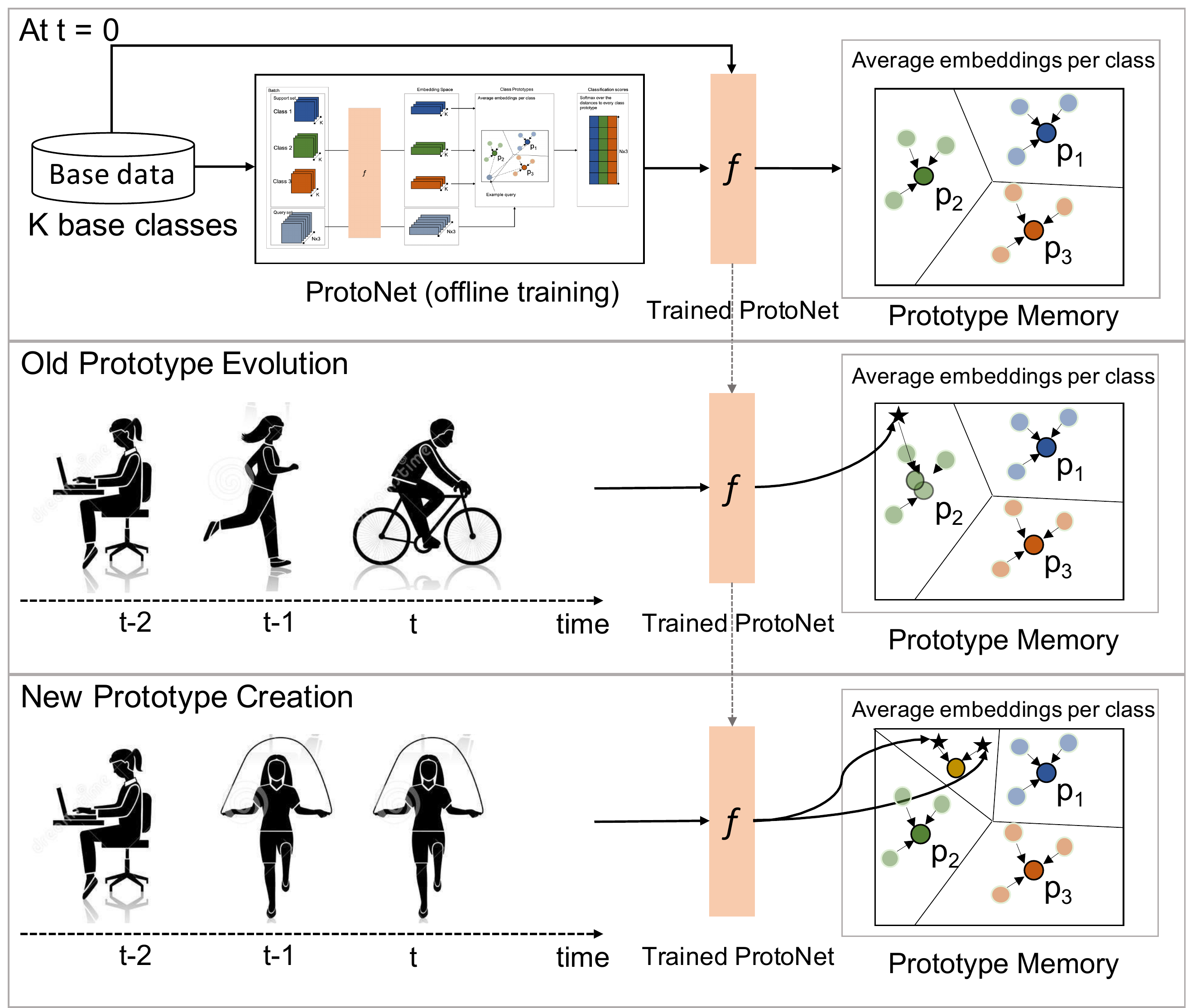}
        \caption{}
        \label{fig:ProtoNEt_online}
    \end{subfigure}
    \hspace{0.01\linewidth}
    \begin{subfigure}[b]{0.49\linewidth}
    \centering
        \includegraphics[width=\columnwidth]{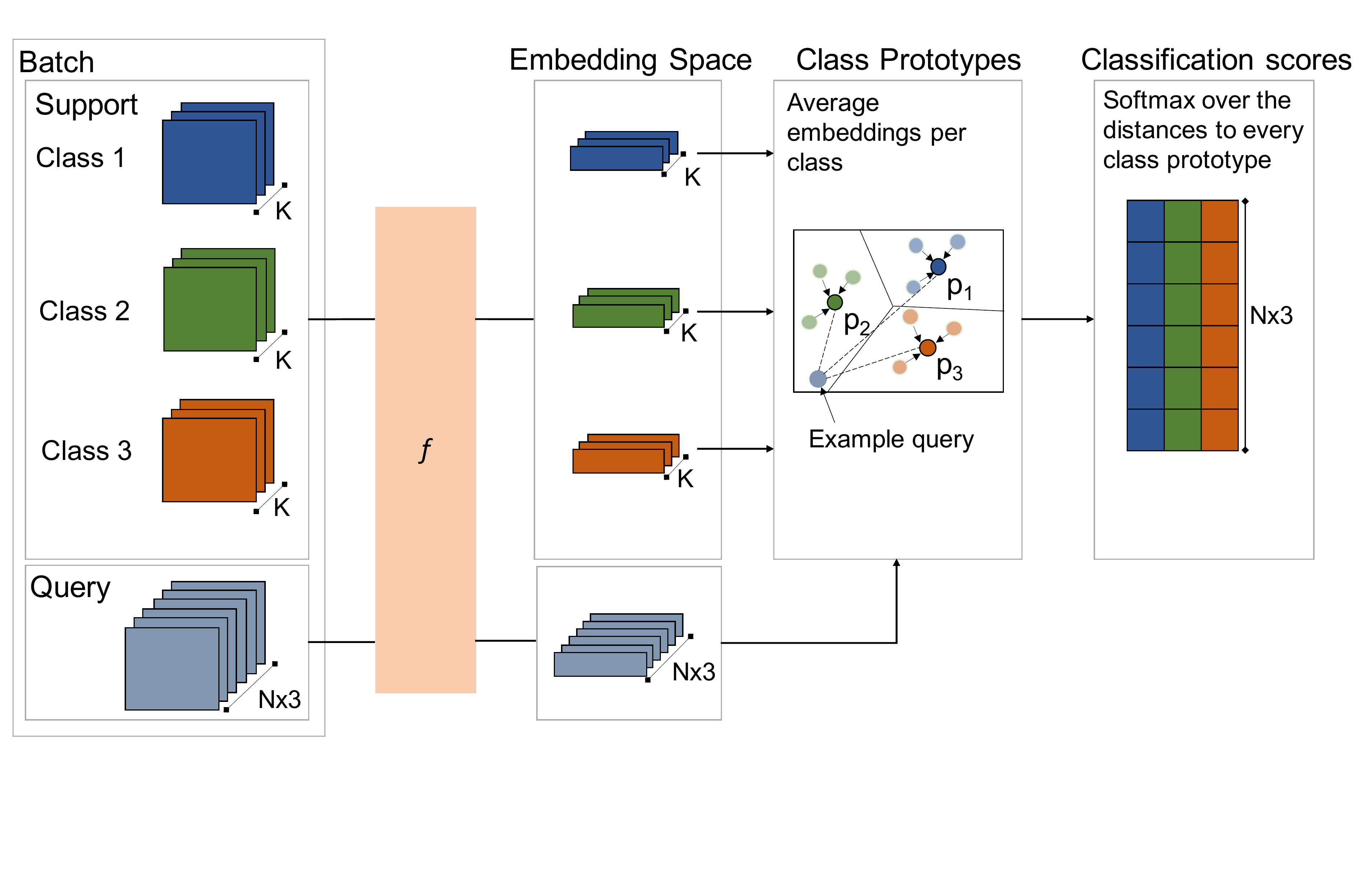}
        \caption{}
        \label{fig:ProtoNEt}
    \end{subfigure}
    \caption{(a) Incremental prototype memory refinement: old prototype evolution using online averaging and new prototype creation when encountering a new activity. (b) Offline Pretraining of Prototypical Network using standard few-shot learning process.}
    \label{fig:ProtoNet_offline_online}
\end{figure}

\subsection{Prototypical Network with Prototype Memory}

First, an evident requirement of our continual learning approach is the ability to keep track of classes learned over time without having to expand the network architecture. To that end, in order to form a knowledge base memory that keeps track of previously learned classes as well as enabled learning from limited number of samples, we leveraged Prototypical Networks \cite{snell2017prototypical} adapted for incremental continual learning (Figure \ref{fig:ProtoNet_offline_online}). This type of network learns a metric space prototype representation for each class. It is adapted to an online setting by storing previously learned class prototypes in a prototype memory. The prototype memory serves as a knowledge memory of previously seen classes that are each represented by a prototype vector corresponding to the average representation of all support samples of the corresponding class. 

\subsubsection{Offline Pretraining}
\label{sec:offline_pretraining}
The first step of our framework is to create a base model able to classify a set of base classes. More specifically, in our continual learning scenario, we assume at time $t=0$ that a subset of classes $C_{base}$ and data samples $D_0$ are available to train a base network $f_{\theta_0}$ following the traditional few-shot learning process (Figure \ref{fig:ProtoNEt}). 

To re-iterate, a prototypical network constructs an embedding space wherein samples cluster around a prototype representation for each class. Training is done episodically wherein, in each episode, a set of classes is provided with a few labeled samples per class. This forms the \textit{support} set $S$ used to form the prototype $p[k]$ of each class $k$ which is defined as the average of all samples belonging to that class:
\begin{equation}
    p[k]=\frac{1}{|S_k|}\sum_{x \in S_k} f_{\theta_0}(x)
    \label{eq:average_proto}
\end{equation}
Given the prototypes of each class, another unlabeled \textit{query} set is used for evaluation. The prototypical network produces a distribution over the classes in memory for the current query sample $f_{\theta_0}(x)$ based on a softmax over the distances to the class prototypes in the embedding space (Equation \ref{eq:euclidan_protoNet}). The training is then performed by minimizing the negative log-likelihood probability of assigning the correct class. 
\begin{equation}
    \hat{y}_{k} = softmax(-\norm{f_{\theta_0}(x)-p[k]}^2)
    \label{eq:euclidan_protoNet}
\end{equation}   

It is important to note that in offline pretraining at every training episode, a support set and query set are randomly sampled every time from the current batch. While this has inspired several meta-learning algorithms, this setup does not allow classes learnt in one episode to be carried forward to the next, which is critical for continual incremental learning. Thus, we leverage an incremental learning paradigm where classes learnt are stored in a \textit{Prototype Memory} ($M_p$) \cite{50004}. As such, after training our base model $f_{\theta_0}$ on $D_0$, we form a base prototype memory $M_p$ that holds base class prototypes computed using all the data samples of each class in $D_0$. More specifically, all data samples in $D_0$ belonging to a class $k$ are passed through the pretrained model $f_{\theta_0}$, and their average is computed to form the corresponding class prototype that is stored in $M_p$. This is depicted in the top part of Figure \ref{fig:ProtoNEt_online}.

\subsubsection{Incremental Prototype Memory Refinement}
\label{sec:online_finetune}
At this point, we have a base model $f_{\theta_0}$ trained to classify the base classes $C_{base}$ for which a corresponding prototype is stored in $M_p$. Now, while the prototype memory allows keeping track of learnt classes, when switching to processing data streams in an online fashion, another obvious requirement is to keep the prototype memory up-to-date by progressively refining it with new incoming data. At every time step \textit{t}, we assume we already have stored a few prototypes corresponding to previously encountered classes in the memory represented by $p_{t}[k]$ where $k$ denotes the corresponding class. With class prototypes estimating the average representation of a class across all support samples, they can be incrementally updated at every time step using online averaging. This can be considered similar to incremental real-time clustering. Given input features $f_{\theta_t}(X)$ for input batch $X$ at time $t$, the prototype is updated using the following:
\begin{equation}
    p_t[k] = \frac{c_{t-1}[k]}{c_t[k]}p_{t-1}[k] + \frac{f_{\theta_t}(X)\mathbbm{1}{[Y=k]}}{c_t[k]}
    \label{eq:online_averaging}
\end{equation}
where $c_t[k] = c_{t-1}[k] + |f_{\theta_t}(X)\mathbbm{1}{[Y=k]}| $ is a count scalar indicating the number of samples added up to time $t$. This enables classes learnt to be carried forward incrementally to future time steps. For novel classes, their prototypes are created by averaging their corresponding embedded samples in the current batch and storing them in the prototype memory. This is further illustrated in Figure \ref{fig:ProtoNEt_online}.

\begin{figure}[t]
    \centering
    \begin{subfigure}[b]{0.322\linewidth}
        \includegraphics[width=1.0\columnwidth]{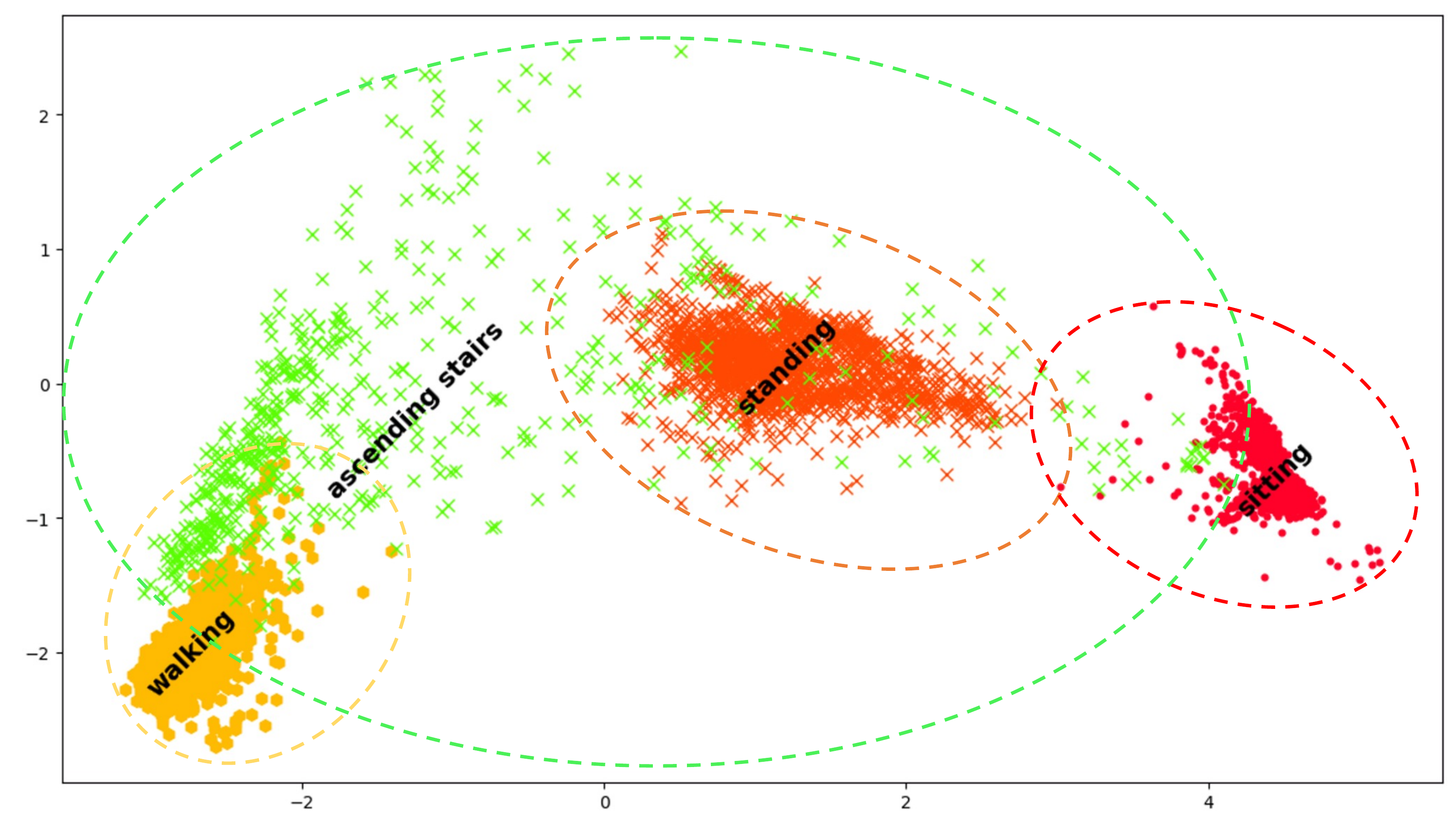}
        \caption{}
        \label{fig:clustering_PAMAP2_offline}
    \end{subfigure}
    \hspace{0.01\linewidth}
    \begin{subfigure}[b]{0.322\linewidth}
        \includegraphics[width=1.0\columnwidth]{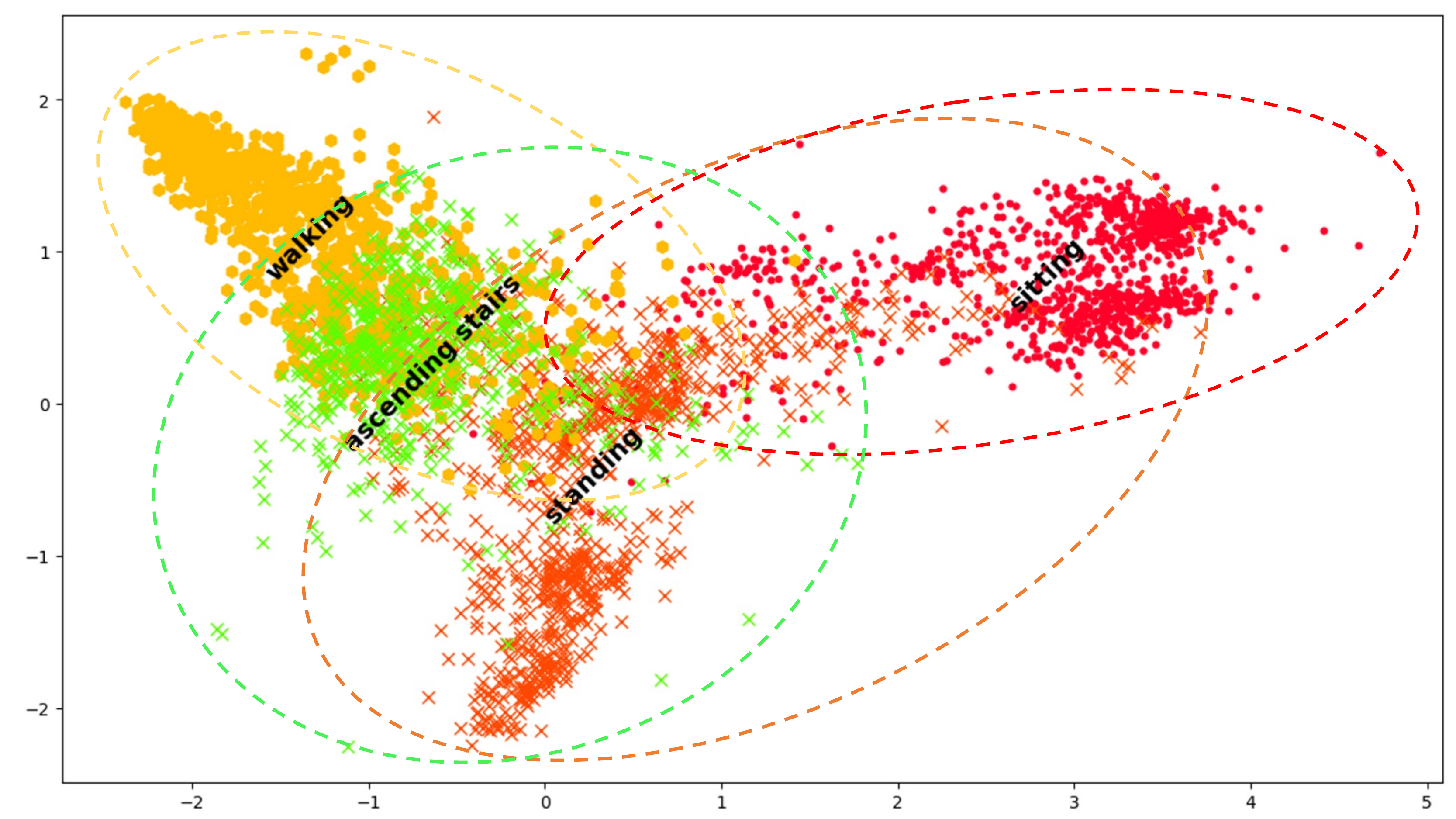}
        \caption{}
        \label{fig:clustering_PAMAP2_finetuning}
    \end{subfigure}
    \begin{subfigure}[b]{0.322\linewidth}
        \includegraphics[width=1.0\columnwidth]{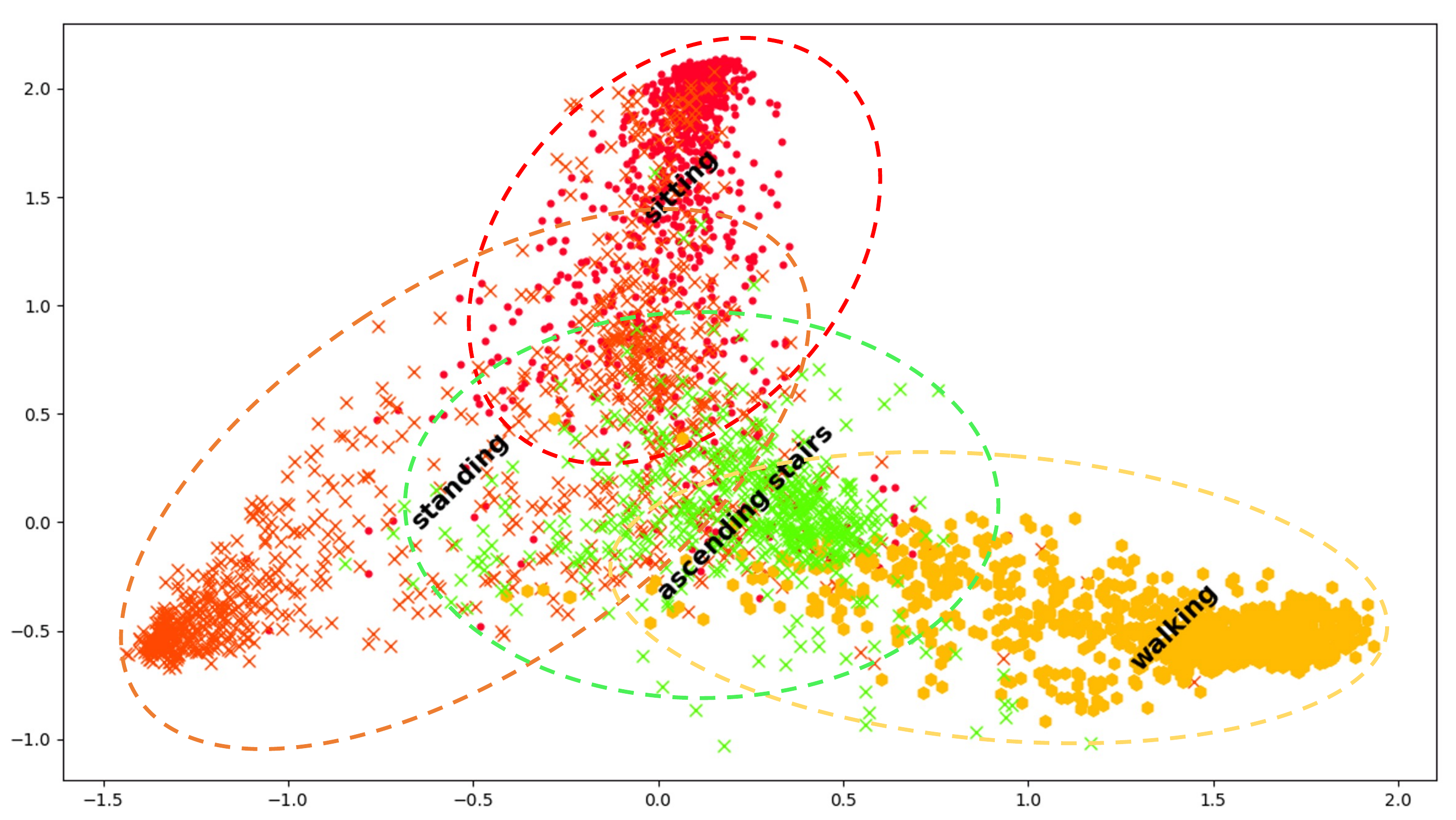}
        \caption{}
        \label{fig:clustering_PAMAP2_LAPNet}
    \end{subfigure}
    \caption{Feature embeddings of base classes --- walking (yellow), sitting (red), standing (orange) --- from PAMAP2 dataset learned during offline pretraining and new unseen class --- ascending stairs (green). (a) After offline pretraining on base classes only: we observe how the feature spatial distribution of new classes is incompatible with the base classes that are well-distinguished and separated. (b) After online finetuning of base and new classes: we observe overlap in the clusters belong to base classes compared to offline pre-training but a narrowing in the feature spatial distribution of the new class. (c) After continual learning of base and new classes using \textit{LAPNet-HAR}: compared to online finetuning, we observe a more compact feature distribution when applying contrastive loss. We still observe some overlap in the class clusters especially for classes that share similar characteristics (such as sitting and standing). We observe some improvement in the new class cluster being separated from the base clusters.} 
    \label{fig:clustering_PAMAP2}
\end{figure}

\subsection{Continual Prototypical Network With Replay Memory}
While prototype memory refinement allows to incrementally update the prototypes with new data and new classes, the feature embedding space initially learned by the base model was trained only on the base classes. Thus when new classes are presented, we observe an incompatible feature space between the new and base classes. This is illustrated in Figure \ref{fig:clustering_PAMAP2_offline} where the base classes are well-separated whereas the feature spatial distribution of the new class is wider and overlaps with the base classes. This poses a challenge in the classification of both new and base classes. It is evident that the feature extractor needs to learn a new joint embedding space that clusters both old and new classes. Updating the model with streaming data using simple online finetuning narrows down the feature spatial distribution of new classes but loses the discriminative clustering of base classes (Figure \ref{fig:clustering_PAMAP2_finetuning}).

In order to solve the above challenges and force the model to learn a joint embedding space for old and new classes in a continual learning process, we proposed to employ a replay memory $M_r$ that holds a subset of previously encountered data samples per class. In our implementation, we applied a random sampling approach of previously observed data, wherein a fixed small subset of samples per class is stored in memory. This replay memory, in turn, is used in the three remaining main components of our \textit{LAPNet-HAR} framework: experience replay, continual replay-based prototype adaptation, and learning using a contrastive loss. The interplay between these components together was shown to improve the model's ability to learn new classes while also mitigating catastrophic forgetting of old classes. In our framework, catastrophic forgetting is incurred at the network parameter level and the prototype representation level. \textit{Experience Replay} addresses forgetting at the network level by retaining observed data distribution, while \textit{replay-based prototype adaptation} addresses forgetting at the stored representation level preventing prototypes from becoming obsolete.
In Figure \ref{fig:clustering_PAMAP2_LAPNet}, we observe some improvement in the clustering of base and new classes when applying \textit{LAPNet-HAR}. It is important to keep in mind that the clustering performance differs for different classes depending on how distinct they are from each other. In Figure \ref{fig:clustering_PAMAP2_LAPNet}, the classes \textit{sitting} and \textit{standing} can share similar characteristics that makes it more challenging to distinguish them apart. In what follows, we present and discuss the utility of each component of \textit{LAPNet-HAR}.

\subsubsection{Experience Replay}

Experience replay presents a well-known method for addressing catastrophic forgetting by mixing exemplar data from previous classes with current data. Re-training the network with previous data helps constrain the network update from becoming biased towards new classes. It also enables the model to better discriminate between both old and new classes. However, this method requires extra memory space for storing exemplar samples per class which can be challenging for on-the-edge applications with limited resources. In our framework, we demonstrate that we do not need to store many exemplar samples in memory to achieve the desired solution. The replay memory $M_r$ is designed to store an equal number of samples per class up to a fixed maximum size (6 samples/class). Moreover, we employ a random sampling approach for selecting the exemplar samples and avoiding computationally expensive herding techniques. At the end of every time step, the replay memory is updated by randomly sampling exemplar samples per class from the current batch and the current data stored in $M_r$. It is also extended when new classes are observed in the current batch. 

\subsubsection{Contrastive Loss}
\label{sec:cont_loss}
As mentioned in Section \ref{sec:offline_pretraining}, the prototypical network $f_\theta$ is optimized to learn a feature embedding space that clusters data samples closer to their corresponding class prototypes. This is done by minimizing the negative log-likelihood. While this enables the network to learn an embedding space in which several points cluster around a single prototype representation for each class, during the continual learning process, the data distribution shifts and the addition of new classes incurs drastic changes in the feature space, as was observed in Figure \ref{fig:clustering_PAMAP2_finetuning}. Therefore, one question that remains, \textit{can we improve the embedding representations learnt to adapt to the changing feature space during continual learning?}

Contrastive loss has been shown effective in learning semantic embeddings by leveraging pairwise relations of data samples \cite{chopra2005learning, hadsell2006dimensionality}. It has been shown to be more effective than cross entropy loss in terms of its robustness to noisy labels and enhancement of class margins. By  leveraging additional semantic similarities between data samples, the model can learn more discriminative embedding representations for each class. This also helps in faster online adaptation when encountering new class samples. Moreover, contrastive loss have been shown to overcome learning capabilities of traditional cross-entropy based classification especially when dealing with small number of training samples \cite{koch2015siamese}. This is particularly useful in our continual learning setting where only a small batch of data is introduced at every time step. There are several formulations of contrastive loss based on metric distance learning or triplets \cite{chopra2005learning,schroff2015facenet} that have been used to learn powerful representations. From our observations in Figure \ref{fig:clustering_PAMAP2_finetuning}, our focus was to push dissimilar samples belonging to different classes (old and new classes) apart while also ensuring similar samples are pulled closer together. Towards that end, we employ a contrastive loss (Equation \ref{eq:ranking_loss}) \cite{hadsell2006dimensionality} that forces the network to narrow down the feature spatial distribution of classes while enlarging the distances among the different classes to be greater than a margin value $m > 0$. It is important to note that, while there have been several recent contrastive loss formulations for different purposes \cite{khosla2020supervised,chen2020simple}, we observed superior performance using the margin-based contrastive loss. Combined with experience replay, the contrastive loss was observed to further enforce inter-class variance between old and new classes. Enforcing inter-class variance reduces interference of new class prototypes on old classes, but at the same time, helps in faster learning of new classes.

Thus, Given a current batch of data $B = D_t \cup M_r$ that includes new samples mixed with replayed samples, similar and different samples are grouped into positive and negative pairs using the ground truth labels. Thus, for a pair of samples $(x_i, y_i)$ and $(x_j,y_j)$, 
\begin{equation}
    \mathcal{L}_{contrastive} = 
    \begin{cases}
    \norm{f_{\theta}(x_i) - f_{\theta}(x_j)}^2 & \textit{if \hspace{1pt} $y_i = y_j$}\\
    \max(0, m - \norm{f_{\theta}(x_i)-f_{\theta}(x_j)}^2) & \textit{if \hspace{1pt} $y_i \neq y_j$}
    \end{cases}
    \label{eq:ranking_loss}
\end{equation}
Finally, the network parameters are updated by minimizing a loss function (Equation \ref{eq:loss_function}) comprised of a softmax-based cross entropy loss encourages samples to be closer to their corresponding prototypes, and a contrastive loss that forces similar samples to be closer to each other (intra-class variance) while dissimilar samples to be at a distance greater than some margin $m$ (inter-class variance).
\begin{equation}
\mathcal{L} = \mathcal{L}_{CE} + \mathcal{L}_{contrastive}
\label{eq:loss_function}
\end{equation}

\subsubsection{Continual Replay-based Prototype Adaptation}
\label{sec:prototype_evolution}

Lastly, with the embedding network evolving with new classes, it is important to prevent the class prototype representations already stored in memory from becoming obsolete, especially for classes not observed in the current batch. This is caused by the changes in data distribution as the model is continually updated which can lead to catastrophic forgetting of old classes. Thus, after the model update and leveraging the replay memory $M_r$, we incorporate a replay-based approach to provide additional information about the current updated state of the embedding space and improving the class prototype representations. More specifically, we adapt a weighted update of previous $p_{t}[k]$ prototypes with replayed exemplars of classes stored in $M_r$ which enables a stabilized prototype optimization. Given $(X_r, Y_r)$ from the replay buffer $M_r$ and network parameters $f_{\theta_{t}}$ at time $t$, the prototype $p_t[k]$ for class $k$ is updated using the following:
\begin{equation}
    p_t[k] = \alpha p_{t}[k] + (1-\alpha)\frac{f_{\theta_{t}}(X_r)\mathbbm{1}{[Y_r=k]}}{|f_{\theta_{t}}(X_r)\mathbbm{1}{[Y_r=k]}|}
    \label{eq:weighted_update}
\end{equation}
where $\alpha$ is a refresh ratio that acts as a learning rate that controls the amount of update applied to the prototypes to incorporate the new latent space while preserving information learned from the previous latent space. Equation \ref{eq:weighted_update} can be seen as equivalent to Equation \ref{eq:online_averaging} with $\alpha = \frac{c_{t-1}[k]}{c_{t-1}[k] + |h_t\mathbbm{1}{y_t=k}|}$. The difference between both equations is that $\alpha \in [0,1]$ is a fixed hyperparameter in Equation \ref{eq:weighted_update}. This helps in stabilizing the prototype trajectory in the ever-evolving latent space. In Section \ref{sec:refresh_ratio}, we discuss and empirically analyze the effect of $\alpha$ on the continual learning process. 

\section{Experiments}

The main objective the following experiments is to asses the effectiveness of our \textit{LAPNet-HAR} framework in balancing the model's plasticity and stability with incremental streaming data of new as well as old classes. In this section, we detail the datasets used, the network architecture of the feature extractor, and the evaluation process. Source code for \textit{LAPNet-HAR} and all the experiments is made available to the community at URL\footnote{\url{https://to-be-added-later.com}}. 

\begin{table*}[t]
\caption{Summary of Selected Datasets}
\label{tab:datasets}
\centering
\begin{tabular}{@{}llccc@{}} \toprule
 \textbf{Datasets} & \textbf{Activity Type} & \textbf{\# of Sensor Channels} & \textbf{\# of Classes} & \textbf{Balanced}  \\
  \midrule
 Opportunity \cite{Chavarriaga:2013:OCB:2537174.2537409} & Daily Gestures & 113 & 17 & \xmark\\
 PAMAP2 \cite{Reiss:2012:CBN:2413097.2413148} & Physical Activities & 52 & 12 & \xmark\\
 DSADS \cite{altun2010comparative} & Daily \& Sports Activities & 45 & 19 & \cmark\\
 Skoda \cite{10.1145/2345770.2345781} & Car Maintenance Gestures & 30 & 10 & \xmark\\
 HAPT \cite{reyes2014human} & Daily Activities \& Postural Transitions & 6 & 12 & \xmark\\
\bottomrule
\end{tabular}

\end{table*}

\subsection{Datasets}
To present a comprehensive analysis of our proposed framework, we assess its performance on 5 widely used publicly available HAR datasets. Table \ref{tab:datasets} summarises the main characteristics of each dataset. 

\subsubsection{Opportunity Dataset}
The Opportunity dataset \cite{Chavarriaga:2013:OCB:2537174.2537409} is a public state-of-the-art HAR dataset. It comprises the readings of motion sensors recorded while 4 subjects executed typical daily activities. From the Opportunity challenge, we replicated the preprocessing steps and train/test split for recognizing 17 different gestures using 113 sensor channels. We applied a sliding window of 800ms with 400ms step size.

\subsubsection{Physical Activity Monitoring Dataset (PAMAP2)}
The PAMAP2 Physical Activity Monitoring dataset \cite{Reiss:2012:CBN:2413097.2413148} contains 18 different physical activities captured using 3 inertial measurement units and a heart rate monitor from 9 subjects. Our implementation modeled 12 activity classes as defined in \cite{Reiss:2012:CBN:2413097.2413148}. We split the dataset into training and testing by setting the data from participants 5 and 6 as the test data while remaining was used for training. We applied a sliding window of 1s with 500ms step size. 

\subsubsection{Daily and Sports Activities Dataset (DSADS)}
The DSADS dataset \cite{altun2010comparative} comprises motion sensor data of 19 daily and sports activities performed by 8 subjects. Each activity was captured for 5 minutes at 25 Hz sampling frequency. The 5-min signals for each activity were divided into 5-sec segments. The tri-axial acceleration, gyroscope, and magnetometer data was captured using 5 MTx orientation trackers placed on the torso, right arm, left arm, right leg, and left leg. The dataset was split into training and testing with the data from subject 7 and 8 used as test data.

\subsubsection{Skoda Dataset}
The skoda dataset \cite{10.1145/2345770.2345781} contains data for 10 quality checks performed by assembly-line workers in a car maintenance scenario. The data was recorded using 10 body-worn accelerometers at 98 Hz sampling rate. The training set consists of 90\% of data from each class, while the remaining 10\% formed the test data. A 1-sec sliding window is used to segment the data with 50\% overlap. 

\subsubsection{Human Activities and Postural Transitions Dataset (HAPT)}
The HAPT dataset \cite{reyes2014human} contains 12 daily activities collected from 30 subjects with a smartphone on their waist. The activities include basic static and dynamic activities such as standing, sitting, walking, etc. as well as postural transitions between static postures such as stand-to-sit, sit-to-stand, etc.. Data captured 3-axial linear acceleration and 3-axial angular velocity at a constant rate of 50 Hz. A 2.56-sec sliding window with 50\% overlap was used to segment the data. The dataset was split into training and testing with the data from participants 29 and 30 used as test data. 

\subsection{Task-Free Data Incremental Protocol}
\label{sec:protocol}
Data incremental learning is a general paradigm that, unlike task-incremental learning, makes no assumption on the data stream. Thus, it provides a more practical and naturalistic scenario for continual learning systems since, at every time step, non-iid data samples are encountered that may contain activity classes already observed before as well as completely new classes. Thus, in addition to addressing the catastrophic forgetting problem, we also consider concept drift \cite{gama2014survey} for previously learned classes due to the fact that data distribution in real life application dynamically change over time. 

The goal is to continue to learn from new data streams starting from a base pretrained model trained on a set of base classes. We propose the following protocol: for a multi-class HAR dataset, classes are randomly split into a set of base classes and new classes. For each base class, the training data is further split into base training data used for offline pre-training and new training data used for streaming. After every time step, each new batch of data is used only once for training. Moreover, the updated model is evaluated on (1) test data containing the base classes only and (2) test data containing new classes already seen so far. In addition to evaluating the model's performance on new classes and base classes separately, we also measure the overall test accuracy on all classes seen so far (base and new combined).

In our experiments, we simulate a data stream where classes are presented randomly in a batch at every time step. We observed that simply applying random sampling over all classes sometimes resulted in a batch containing all classes and 1 sample per class, especially given the small batch size (e.g. 20 samples). Thus, in order to ensure a more practical streaming behavior, we restricted a batch to include at most 5 classes from remaining old and new classes combined.

\begin{table*}[t]
\caption{Hyperparameter Configuration}
\label{table:hyperparameter}
\centering
\ra{1.4}
\begin{tabular}{@{}clccccc@{}}\toprule
& & \textbf{Opportunity} & \textbf{PAMAP2} & \textbf{DSADS} & \textbf{Skoda} & \textbf{HAPT} \\
\cmidrule{1-7}
\multirow{5}{*}{\rotatebox[origin=c]{90}{Offline Pretraining}}
& Batch size & 200 & 200 & 200 & 200 & 200 \\
& \# Base classes & 5 & 5 & 5 & 5 & 5 \\
& \% Training data / class & 50\% & 50\% & 50\% & 50\% & 50\% \\
& Epochs & 100 & 100 & 100 & 100 & 100 \\
& Learning rate & 0.001 & 0.001 & 0.001 & 0.001 & 0.001\\
\cmidrule{1-7}
\multirow{5}{*}{\rotatebox[origin=c]{90}{Continual learning}} 
& Batch size & 20 & 20 & 20 & 20 & 20 \\
& Replay buffer size / class & 6 & 6 & 6 & 6 & 6 \\
& Contrastive margin ($m$) & 3 & 2 & 1 & 1 & 1 \\
& Refresh ratio ($\alpha$) & 1.0 & 0.2 & 0.8 & 0.5 & 0.5\\
& Learning rate & 0.001 & 0.001 & 0.001 & 0.001 & 0.001 \\

\bottomrule
\end{tabular}

\end{table*}

\subsection{Architecture and Training Configuration}

While the network $f_\theta$ can be any type of architecture, in our experiments we apply the same architecture across all datasets in order to maintain a fair comparison. DeepConvLSTM \cite{ordonez2016deep}, which has been widely used for HAR datasets, uses a combination of convolutional and recurrent layers. Raw sensor signals are processed by four 1D convolutional layers, of 64 filters with kernel size $1\times5$, which extract feature maps from the wearable data. Then, the feature maps are passed to two LSTM layers with 128 hidden units for sequential processing. The original DeepConvLSTM network includes a softmax classification layer at the end. However, in our prototypical network, the softmax layer is removed as the model is trained to learn a metric space in which classification can be performed by computing distances to the prototype representations of each class. 

While maintaining a fair comparison premise across datasets by using the same network architecture, we empirically optimize hyperparameter configurations for each dataset to achieve the highest performance. Table \ref{table:hyperparameter} lists the optimised hyperparameters used for both pre-training and continual learning processes in our \textit{LAPNet-HAR} framework. We discuss the effect of tuning some hyperparameters on the learning process in Section \ref{sec:discussion}.

\subsection{Evaluation Metrics}

In order to evaluate \textit{LAPNet-HAR}'s ability to continually learn and adapt to new classes while retaining knowledge about previous classes, we compute several measures on the held-out test data.  

\subsubsection{Performance Measures}

Using the held-out test data as explained in Section \ref{sec:protocol}, we compute three performance measures, more specifically using average F1-score (Equation \ref{eq:f1}): (1) \textit{Base} performance which measures the model's ability to recognize the base classes $C_{base}$ it was pretrained on, (2) \textit{New} performance which measures the model's ability to recognize the new classes seen so far, and (3) \textit{Overall} performance which measures the overall ability of the model to recognize both base and new classes seen so far. The base performance will indicate the model's stability while the new performance describes the plasticity of the model.  
\begin{equation}
    F1\textnormal{-}score = \frac{2}{C}\sum_{i=1}^{C}\frac{prec_i\times recall_i}{prec_i+recall_i}
    \label{eq:f1}
\end{equation}
\subsubsection{Forgetting Measure}
\textit{Forgetting} of a given task is defined as the difference between the maximum knowledge gained by the model about the task throughout the past learning process and the current knowledge the model has \cite{chaudhry2018riemannian}. This metric helps provide a quantitative measure to better understand the model's stability and ability to retain knowledge about previous tasks. 

While this measure was defined for a task-incremental learning scenario, we leverage the same formulation but adapt a modified forgetting measure that can be suitable for our data-incremental learning protocol. More specifically, in our learning setup, we do not deal with incrementally adding tasks but rather incrementally adding data that can include new classes or new observations of old classes. Thus, after the model is trained incrementally till time step $t$, the forgetting score $f_t^j$ is computed as the difference between the maximum performance gained for class $j$ throughout the learning process up to $t-1$ and the current performance for class $j$ at time $t$. Adapting the normalized version introduced by \cite{JHA2021}, the forgetting measure is defined as:
\begin{equation}
    f_t^j = 1 - \frac{a_{t,j}}{\underset{l \in 1,...,t-1}{\max} a_{l,j}}
\end{equation}
where $a_{t,j}$ is the performance measure for class $j$ at current time step $t$. We compute the forgetting measure for each of the base classes the model was pretrained on. Then, we take the average forgetting measure across all base classes,
\begin{equation}
F_t = \frac{1}{|C_{base}|} \sum_{j \in C_{base}} f_t^j
\label{eq:mean_forgetting}
\end{equation}
A high forgetting measure indicates severe catastrophic forgetting of base classes. 

\subsubsection{Intransigence Measure}
While the forgetting measure is crucial for understanding the catastrophic forgetting effect observed in continual learning systems, another key aspect that also needs to be analyzed is the model's ability to learn new classes. This indicates the plasticity of the model and typically interplays with the model's forgetting of old classes. Thus, analyzing both measures together provides a better understanding of our framework's effectiveness in continual learning. Again, we leverage the \textit{intransigence} measure defined by Chaudhry \textit{et al.}
 \cite{chaudhry2018riemannian} that is defined as the inability of a model to learn new tasks.
 
Similar to our forgetting measure, we modify the intransigence measure to fit out learning setup by considering time steps instead of tasks. Therefore, we compare the performance measure of the model ($a_{t,j}$) at time $t$ on new class $j$ to a reference model's performance ($a_j^*$) on the same class. We define the reference model to be the prototypical network trained offline on all the training data that contains all classes. The reference model is basically the standard classification model that serves as the upper bound of the continual learning performance. Therefore, at time $t$, the intransigence measure for class $j$ is, 
\begin{equation}
I_t^j = a_j^* - a_{t,j}
\end{equation}
Similar to Equation \ref{eq:mean_forgetting}, we take the mean across the intransigence measures of all new classes excluding base classes. A model with lower intransigence score indicates better ability to learn new classes.  

\subsection{Baselines}
\label{sec:baseline}

We compare our framework against two baselines: (1) \textbf{offline supervised} which implements a standard offline training of our prototypical network on all training dataset and (2) \textbf{online finetuning} which applies a simple update of the model with cross entropy loss using only the data as it is streamed without experience replay, prototype adaptation, and contrastive loss. Online finetuning is essentially the prototypical network with prototype memory, where the prototypes are updated online using online averaging as explained in Section \ref{sec:online_finetune}. Establishing these two conditions as baselines allows us to assess how our method performs in relation to these two established measures. Thus, the offline supervised baseline serves as an upper bound while online finetuning baseline serves as a lower bound.

\begin{table*}[t]
\caption{Summary Results. \textsuperscript{\ddag}: without experience replay and contrastive loss, \textsuperscript{\dag}: without contrastive loss.}
\label{table:results}
\begin{subtable}{\columnwidth}
\centering
\caption{Base Performance}
\label{tab:base}
\begin{tabular}{@{}lccccc@{}}\toprule
\textbf{Models} & \textbf{PAMAP2} & \textbf{HAPT} & \textbf{Opportunity} & \textbf{DSADS} & \textbf{Skoda} \\
\midrule
Offline & 80.56 $\pm$ 5.84 & 75.62 $\pm$ 4.53 & 62.68 $\pm$ 2.23 & 66.85 $\pm$ 10.86 & 76.46 $\pm$ 14.13\\
\midrule \midrule
LAPNet-HAR\textsuperscript{\ddag} & 46.94 $\pm$ 19.64 & 23.02 $\pm$ 17.24 & 14.21 $\pm$ 9.20 & 34.64 $\pm$ 22.21 & 42.48 $\pm$ 14.59 \\
LAPNet-HAR\textsuperscript{\dag} & 63.29 $\pm$ 12.86 & 42.92 $\pm$ 19.78 & 31.07 $\pm$ 7.98 & 49.57 $\pm$ 14.93 & 65.05 $\pm$ 13.90 \\
LAPNet-HAR & \textbf{69.66 $\pm$ 9.11} & \textbf{48.73 $\pm$ 10.87} & \textbf{35.47 $\pm$ 11.46} & \textbf{51.84 $\pm$ 22.27} & \textbf{74.28 $\pm$ 4.36} \\
\midrule \midrule
Online finetuning & 34.04 $\pm$ 30.80 & 20.02 $\pm$ 15.16 & 19.73 $\pm$ 13.01 & 26.43 $\pm$ 21.79 & 35.19 $\pm$ 12.78 \\
\bottomrule
\\
\end{tabular}
\end{subtable}

\begin{subtable}{\columnwidth}
\centering
\caption{New Performance}
\label{tab:new}
\begin{tabular}{@{}lccccc@{}}\toprule
\textbf{Models} & \textbf{PAMAP2} & \textbf{HAPT} & \textbf{Opportunity} & \textbf{DSADS} & \textbf{Skoda} \\
\midrule
Offline & 83.33 $\pm$ 2.09 & 67.69 $\pm$ 3.46 & 55.92 $\pm$ 3.40 & 64.70 $\pm$ 3.67 & 89.04 $\pm$ 10.69 \\
\midrule \midrule
LAPNet-HAR\textsuperscript{\ddag} & 54.17 $\pm$ 6.07 & 14.68 $\pm$ 10.13 & 25.33 $\pm$ 4.35 & 35.27 $\pm$ 14.72 & 56.09 $\pm$ 26.34 \\
LAPNet-HAR\textsuperscript{\dag} & 64.50 $\pm$ 13.54 & 20.57 $\pm$ 4.53 & 23.77 $\pm$ 4.62 & 36.24 $\pm$ 9.28 & 60.04 $\pm$ 26.72 \\
LAPNet-HAR & \textbf{71.70 $\pm$ 6.61} & \textbf{28.28 $\pm$ 6.53} & \textbf{32.94 $\pm$ 4.66} & \textbf{45.63 $\pm$ 12.22} & \textbf{82.14 $\pm$ 15.46} \\
\midrule \midrule
Online finetuning & 51.92 $\pm$ 27.57 & 19.61 $\pm$ 10.99 & 23.82 $\pm$ 2.47 & 31.26 $\pm$ 16.97 & 55.31 $\pm$ 27.17 \\
\bottomrule
\\
\end{tabular}
\end{subtable}

\begin{subtable}{\columnwidth}
\centering
\caption{Overall Performance}
\label{tab:overall}
\begin{tabular}{@{}lccccc@{}}\toprule
\textbf{Models} & \textbf{PAMAP2} & \textbf{HAPT} & \textbf{Opportunity} & \textbf{DSADS} & \textbf{Skoda} \\
\midrule
Offline & 78.33 $\pm$ 1.90 & 64.57 $\pm$ 3.09 & 52.88 $\pm$ 4.29 & 61.61 $\pm$ 4.90 & 80.34 $\pm$ 7.16 \\
\midrule \midrule
LAPNet-HAR\textsuperscript{\ddag} & 45.85 $\pm$ 7.56 & 14.86 $\pm$ 10.97 & 18.81 $\pm$ 2.05 & 31.21 $\pm$ 7.08 & 42.56 $\pm$ 15.79 \\
LAPNet-HAR\textsuperscript{\dag} & 58.76 $\pm$ 6.26 & 26.00 $\pm$ 8.52 & 21.79 $\pm$ 4.74 & 36.45 $\pm$ 6.70 & 57.33 $\pm$ 16.66 \\
LAPNet-HAR & \textbf{65.92 $\pm$ 5.16} & \textbf{32.17 $\pm$ 5.33} & \textbf{28.41 $\pm$ 3.64} & \textbf{43.57 $\pm$ 10.26} & \textbf{74.78 $\pm$ 8.98} \\
\midrule \midrule
Online finetuning & 38.84 $\pm$ 23.83 & 16.62 $\pm$ 11.13 & 19.67 $\pm$ 4.28 & 26.62 $\pm$ 9.10 & 37.46 $\pm$ 14.69 \\

\bottomrule
\end{tabular}
\end{subtable}

\end{table*}

\section{Results}

In this section, we present the experimental results on the 5 HAR datasets. We perform ablation experiments on \textit{experience replay}, \textit{replay-based prototype adaptation}, and \textit{contrastive loss}. The goal of these components in our \textit{LAPNet-HAR} framework is to enable the model to continually learn new classes while mitigating catastrophic forgetting and adapt the feature embedding space to distribution shifts that occur during streaming. Thus, we evaluate the contribution of each component to the overall performance. It is important to note that the boost in the continual learning process is not explained by any single component of \textit{LAPNet-HAR}, but by their composition.

\subsection{Performance Analysis}
Table \ref{table:results} reports the mean and standard deviation of the \textit{Base}, \textit{New}, and \textit{Overall} F1-score for each dataset over 5 runs. From the experimental results, we observe that the combination of \textit{experience replay}, \textit{replay-based prototype adaptation}, and \textit{contrastive loss} significantly improves the recognition performance of new as well as base classes across all datasets. This in turn leads to higher \textit{Overall} performance across all datasets when applying \textit{LAPNet-HAR}. 

As lower bound, the online finetuning baseline suffers the most catastrophic forgetting with \textit{Base} performance dropping by 45\% on average across all datasets compared to the offline upper bound \textit{Base} performance. We observe that incorporating replay-based prototype adaptation (\textit{LAPNet-HAR\textsuperscript{\ddag}}) significantly reduced catastrophic forgetting of base classes compared to online finetuning specifically for PAMAP2, DSADS and Skoda, while performance on new classes was comparable or less than the lower baseline. This demonstrates the benefit of updating and stabilizing the prototype representation progression as the embedding network evolves at every time step. However, the model's ability to learn new classes is still lacking. This can be due to the randomness in the class and data sequence at every time step which results in the scarcity of data samples when a new class is introduced. More specifically, the model does not converge on a new class or task before moving to the next one. While this reflects a more naturalistic learning scenario, this hinders the model's ability to properly separate the classes seen. This is in stark contrast to task-incremental learning where all data samples of a new class or task is observed before moving forward. Employing experience replay to further mitigate catastrophic forgetting (\textit{LAPNet-HAR\textsuperscript{\dag}}) resulted in further increase in \textit{Base} performance for all datasets, improving by a range of 14\% to 23\% compared to \textit{LAPNet-HAR\textsuperscript{\ddag}}. On the other hand, \textit{New} performance slightly improved for HAPT, Skoda, and DSADS, while remained comparable for Opportunity and PAMAP2. In terms of the \textit{Overall} performance, while \textit{LAPNet-HAR\textsuperscript{\dag}} significantly improved compared to online finetuning for all datasets, it was still low relative to the offline upper bound, reaching a minimum gap of 23\%. This indicates the clustering ability of the model is weakened during continual learning of new and old classes. Thus, as mentioned previously, a contrastive loss is employed to improve inter-class separation during learning which is evident in the significant improvement in the \textit{Base}, \textit{New}, and \textit{Overall} performances of \textit{LAPNet-HAR} across all datasets. Skoda achieved an \textit{Overall} performance of 74.78\%, only a 6\% difference compared to the offline performance. For Opportunity and HAPT, on the other hand, we observe lower performance indicating a more challenging continual learning process. This can be attributed to the highly imbalanced nature of the datasets, the larger number of activity classes, as well as the larger similarity and overlap in the classes. For example, the  gestures activities (\textit{OpenDoor1}/\textit{OpenDoor2} or \textit{CloseDoor1}/\textit{CloseDoor2}) in Opportunity and the postural transitions (\textit{sit to stand}/\textit{stand to sit}) in HAPT can lead to overlapping cluster representations due to similarities in the activities.

\begin{figure}[t]
    \centering
    \includegraphics[width=1.\columnwidth]{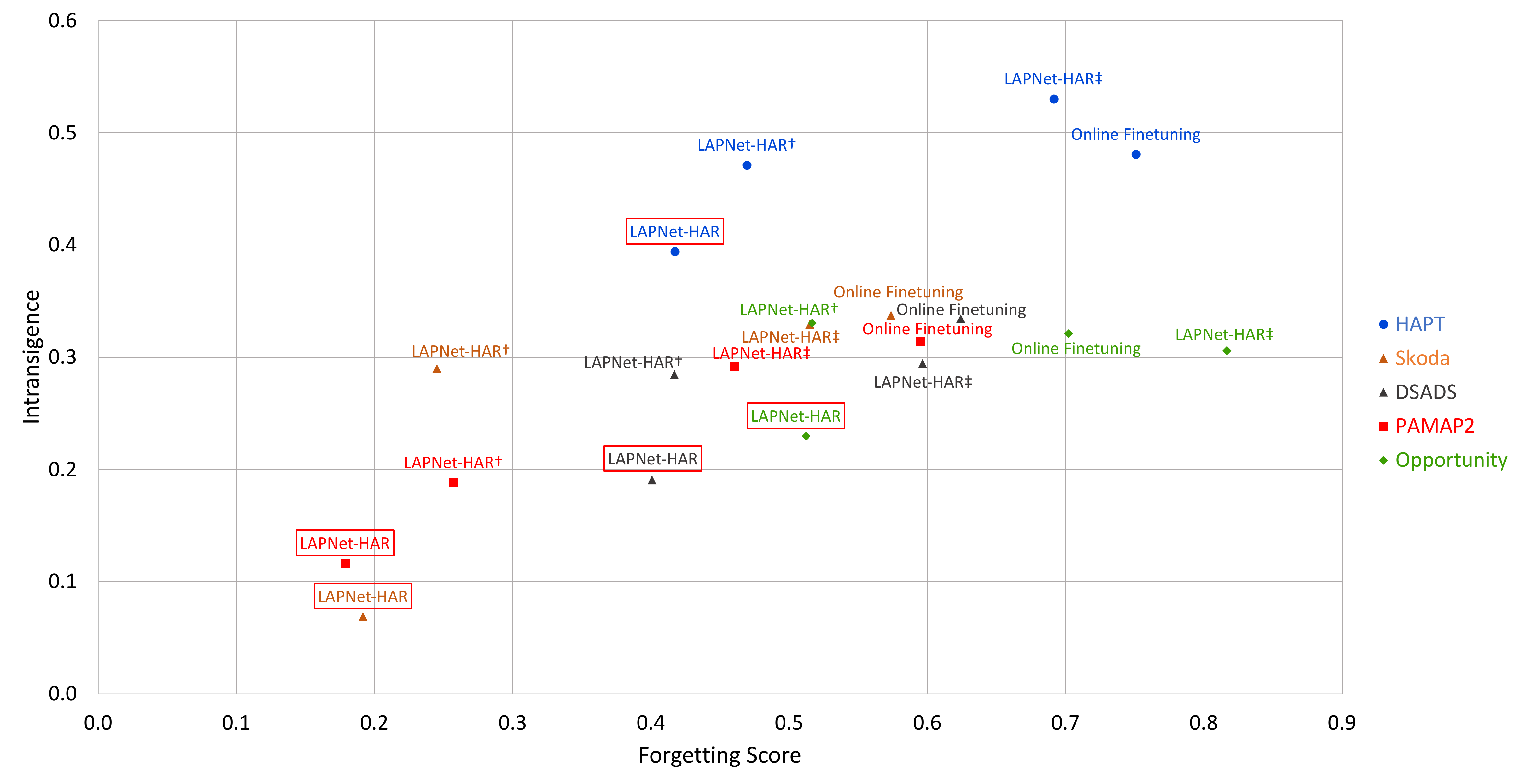}
    \caption{Interplay between forgetting and intransigence. The methods framed in red denote the method with the best trade-off between forgetting and intransigence for each dataset.}
    \label{fig:forgetting_intransigence}
\end{figure}{}

\subsection{Interplay of Forgetting and Intransigence}

To further inspect the catastrophic forgetting and plasticity of our framework, we study the interplay of forgetting and intransigence in Figure \ref{fig:forgetting_intransigence}. To reiterate, there is an inherent trade-off between a model's ability to retain information about old classes and its ability to learn new classes without interfering with previous knowledge. Intuitively, if a model is heavily pushed to preserve past knowledge, it will forget less but have high intransigence. If, in contrast, preserving past information is weak, while intransigence will be low, the model will suffer from catastrophic forgetting.  An ideal continual learning model would have low forgetting and intransigence measures, thus efficiently leveraging the model's learning capacity.

We observe in Figure \ref{fig:forgetting_intransigence} that, across all datasets, \textit{LAPNet-HAR} exhibits the best trade-off between forgetting and intransigence with PAMAP2 and Skoda losing around $18-19\%$ of their best performance across all classes throughout the continual learning process. Focusing on the forgetting measure, Opportunity dataset exhibits a more challenging learning process with forgetting measure reaching $51\%$ when applying \textit{LAPNet-HAR} which still outperforms other methods and the reference baseline. As expected, online finetuning exhibits the highest forgetting reaching $ \sim 60\% $ for PAMAP2, Skoda, and DSADS and exceeding $70\%$ for Opportunity and HAPT datasets. While maintaining low forgetting, \textit{LAPNet-HAR} is also observed to have the lowest intransigence across all datasets indicating that the model is also capable of continually learning new classes. It is evident that forgetting and intransigence of a model can be affected by dataset characteristics, as was also observed in the performance measures, especially with how old and new classes interfere with each other.

\section{Discussion}
\label{sec:discussion}

\subsection{Parameter Configurations}
In this section, we analyze the effect of tuning the (1) contrastive margin and (2) the refresh ratio $\alpha$ on the performance measures and the continual learning process.  

\begin{figure}[t]
    \centering
    \begin{subfigure}[b]{0.32\linewidth}
        \includegraphics[width=1.0\columnwidth]{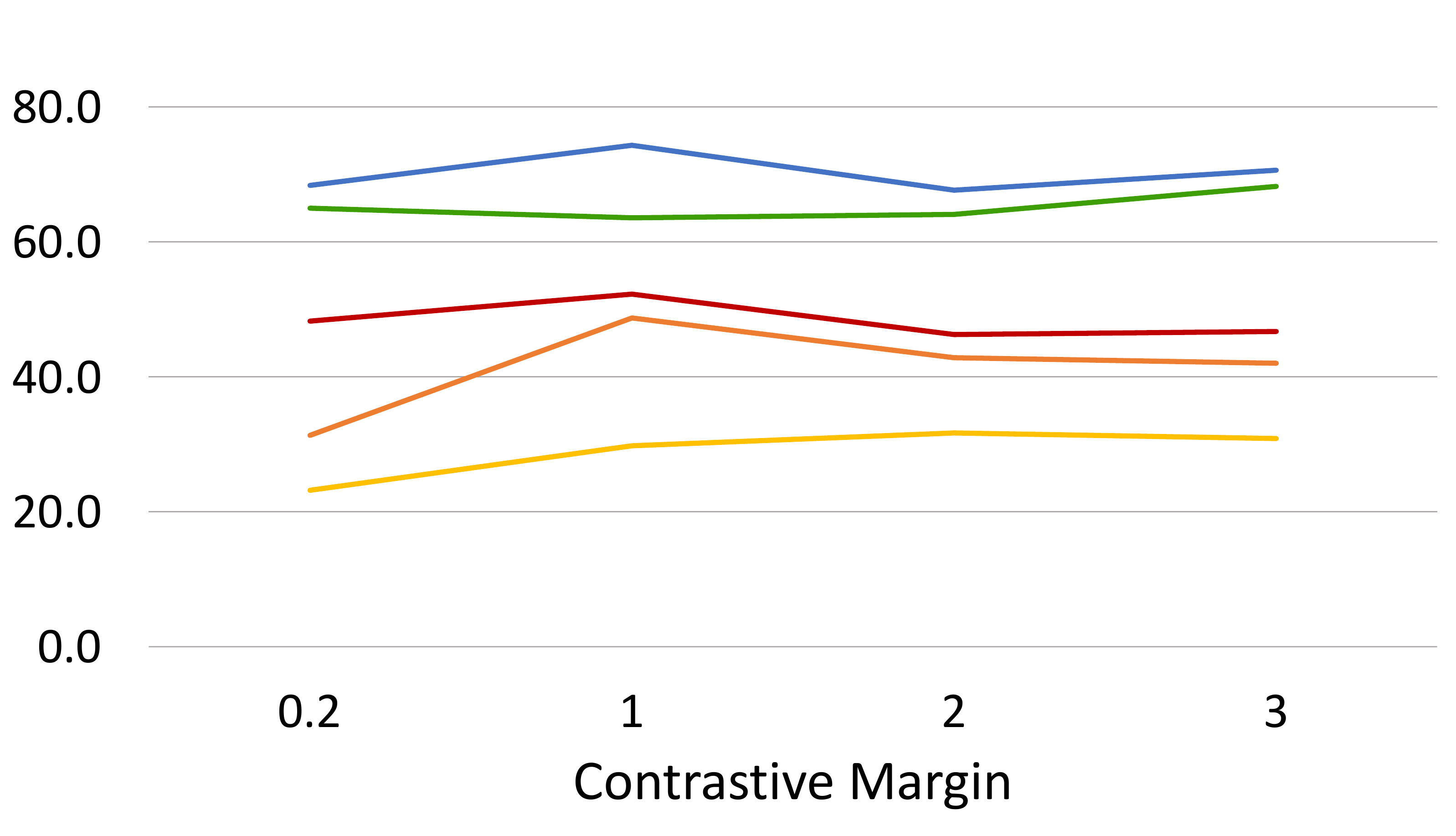}
        \caption{Base Performance}
        \label{fig:vary_margin_base}
    \end{subfigure}
    \begin{subfigure}[b]{0.32\linewidth}
        \includegraphics[width=1.0\columnwidth]{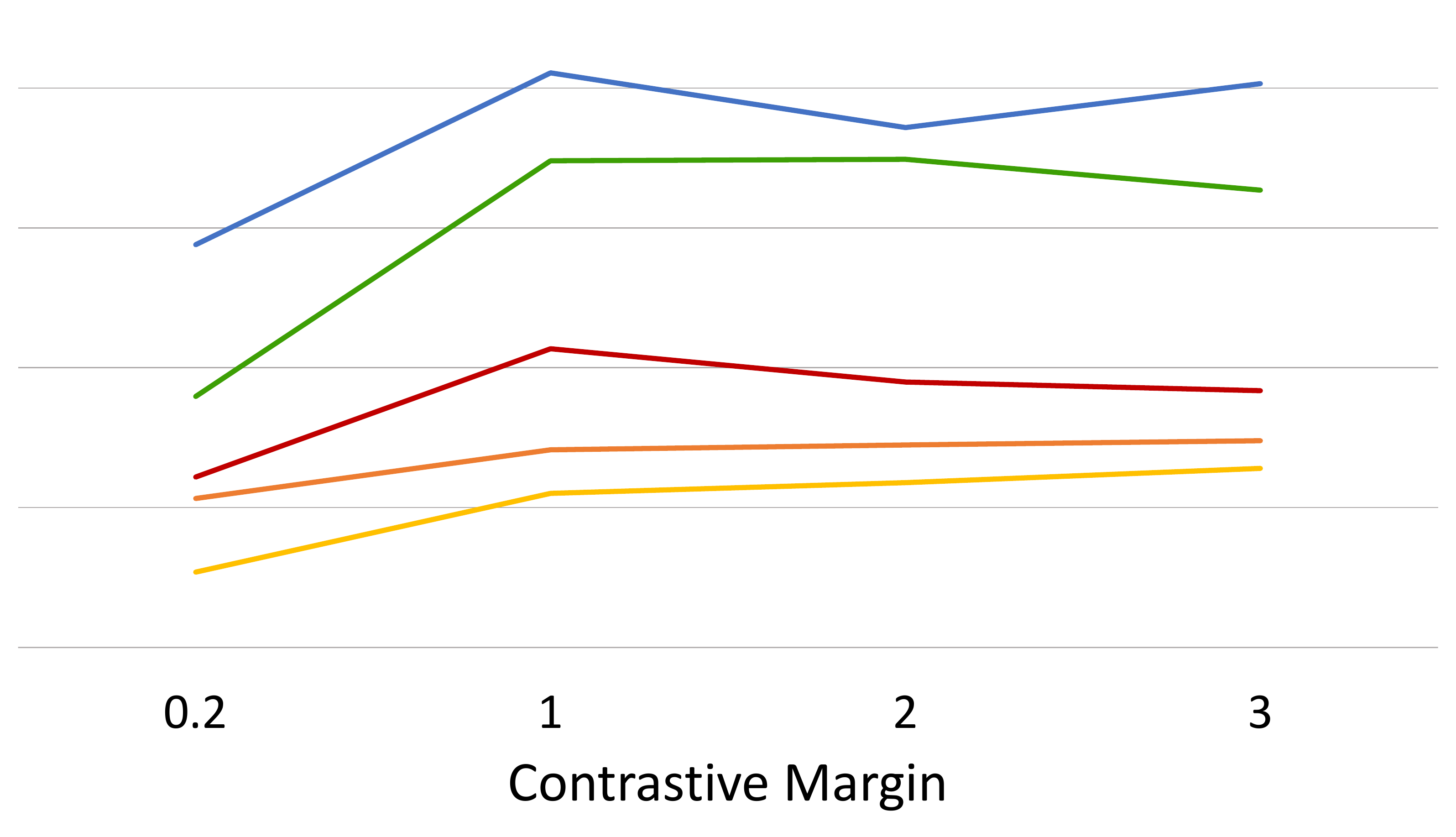}
        \caption{New Performance}
        \label{fig:vary_margin_new}
    \end{subfigure}
    \begin{subfigure}[b]{0.32\linewidth}
        \includegraphics[width=1.0\columnwidth]{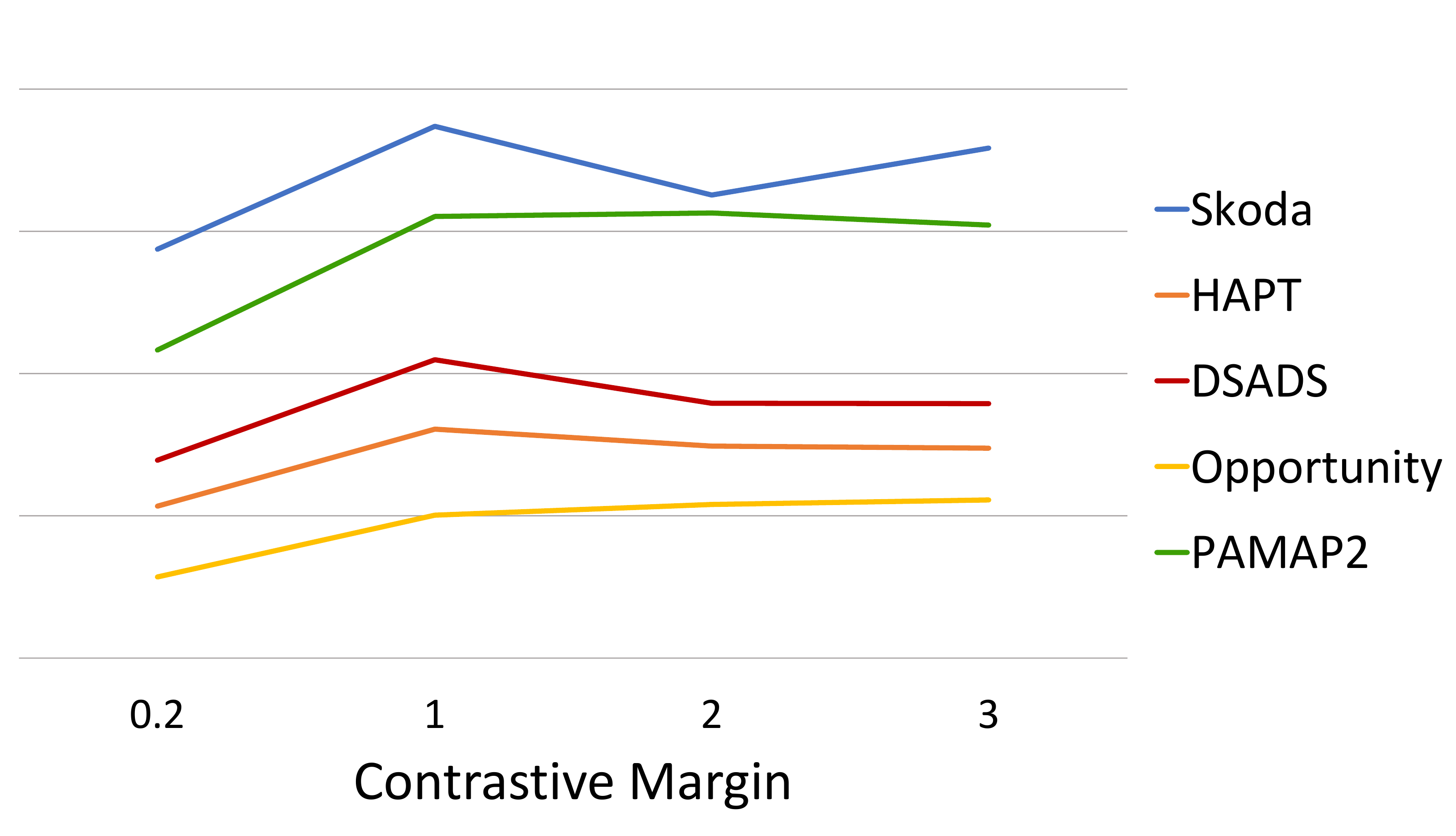}
        \caption{Overall Performance}
    \label{fig:vary_margin_overall}
    \end{subfigure}
    \caption{Effect of contrastive margin on (a) base, (b) new, and (c) overall performance across all datasets.}
    \label{fig:vary_margin}
\end{figure}{}

\subsubsection{Effect of Contrastive Margin}
As explained in Section \ref{sec:cont_loss}, the contrastive loss forces samples that belong to different classes to be farther apart by some predefined margin $m$ enforcing inter-class separation. Keeping all other parameters fixed (Table \ref{table:hyperparameter}) and varying the margin $m \in \{0.2, 1, 2, 3\}$, we analyze its effect on the performance measures across all datasets (Figure \ref{fig:vary_margin}). The range was defined empirically based on observations of the intra-class euclidean distance of embeddings. We observe that the margin has a larger effect on the \textit{New} performance with a significant increase when moving from margin $=0.2$ to $1$ across all datasets. For Skoda, DSADS and PAMAP2, the \textit{New} performance increases by at least 18\% and at most 34\% compared to low margin $m = 0.2$. The \textit{Base} performance for all datasets, except for HAPT and PAMAP2, did not have a significant change ($\sim 5\%$) with the increase in margin. This can be due to the fact that the model is already pre-trained to recognize and separate the base classes, and thus during the continual learning process more emphasis is placed on learning to separate the new classes. Since a small number of random samples are streamed (20 samples), fast adaptation to new classes using only cross entropy is limited and challenging. Thus, incorporating a contrastive loss forces the model to learn an embedding space that separates different classes even with limited samples. Overall, a contrastive margin $m=1$ resulted in a boost in \textit{Overall} performance with a slight increase when further increasing the margin for some datasets.  

\begin{figure}[t]
    \centering
    \begin{subfigure}[b]{0.32\linewidth}
        \includegraphics[width=1.0\columnwidth]{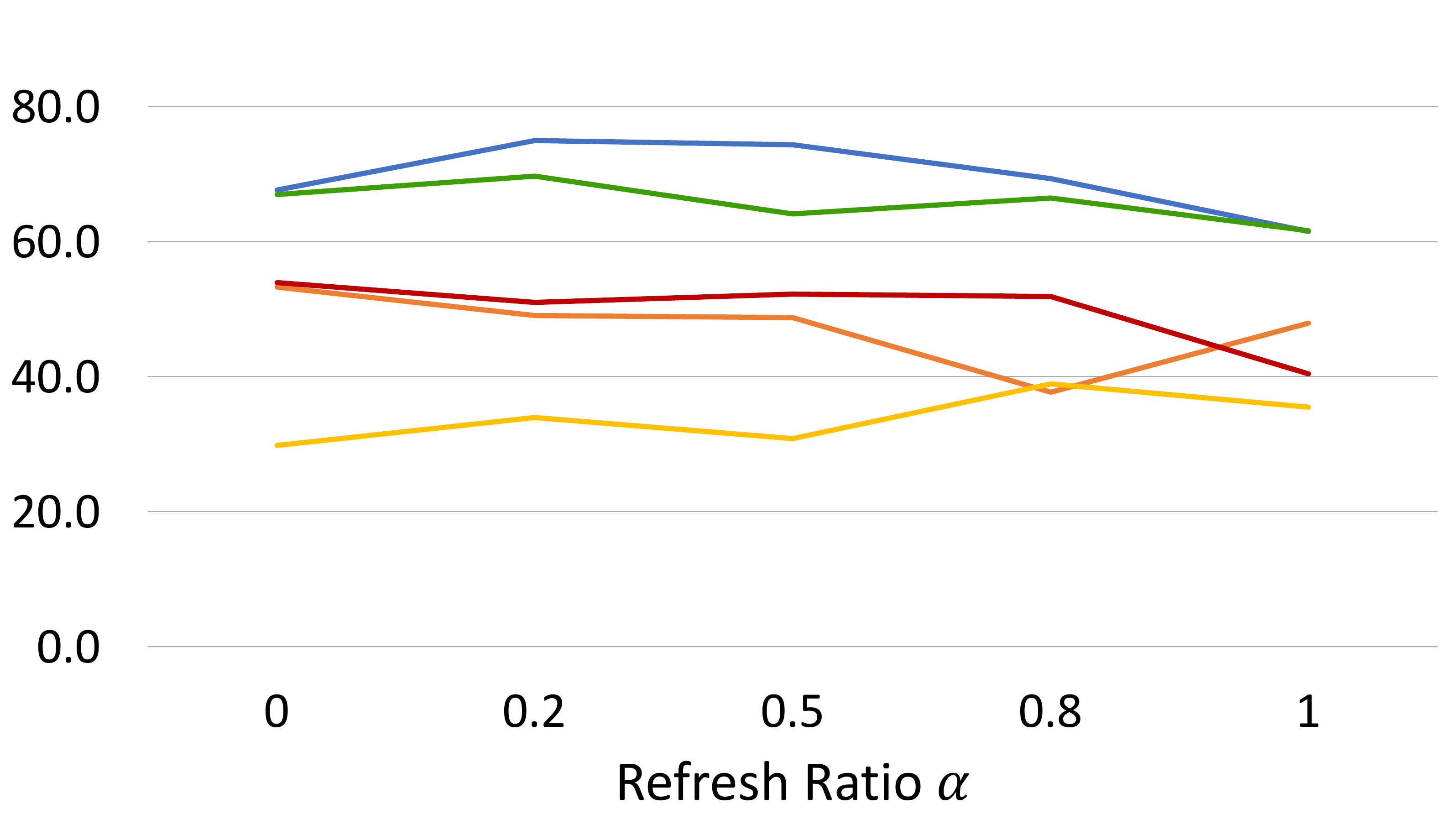}
        \caption{Base Performance}
        \label{fig:vary_alpha_base}
    \end{subfigure}
    \begin{subfigure}[b]{0.32\linewidth}
        \includegraphics[width=1.0\columnwidth]{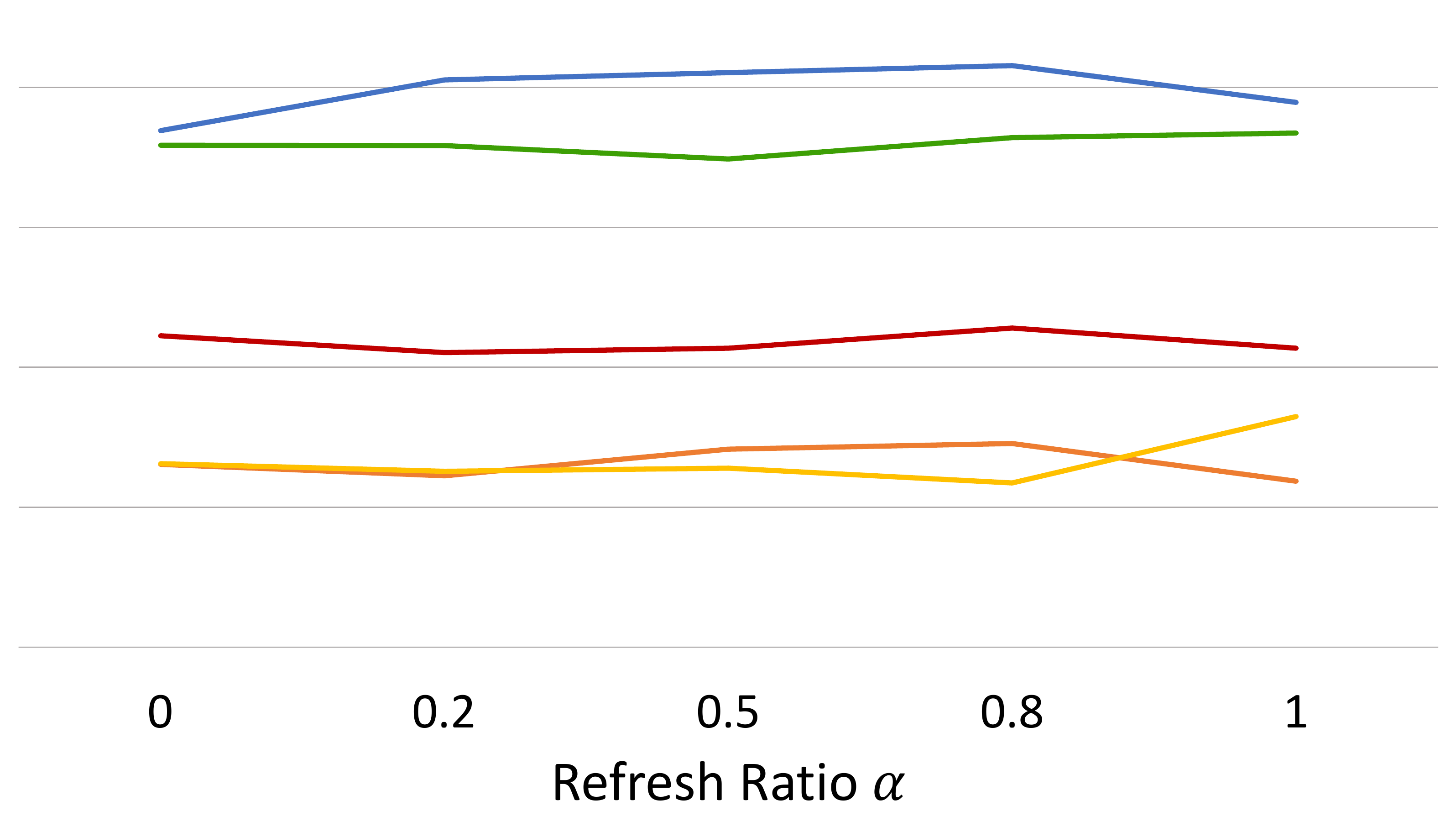}
        \caption{New Performance}
        \label{fig:vary_alpha_new}
    \end{subfigure}
    \begin{subfigure}[b]{0.32\linewidth}
        \includegraphics[width=1.0\columnwidth]{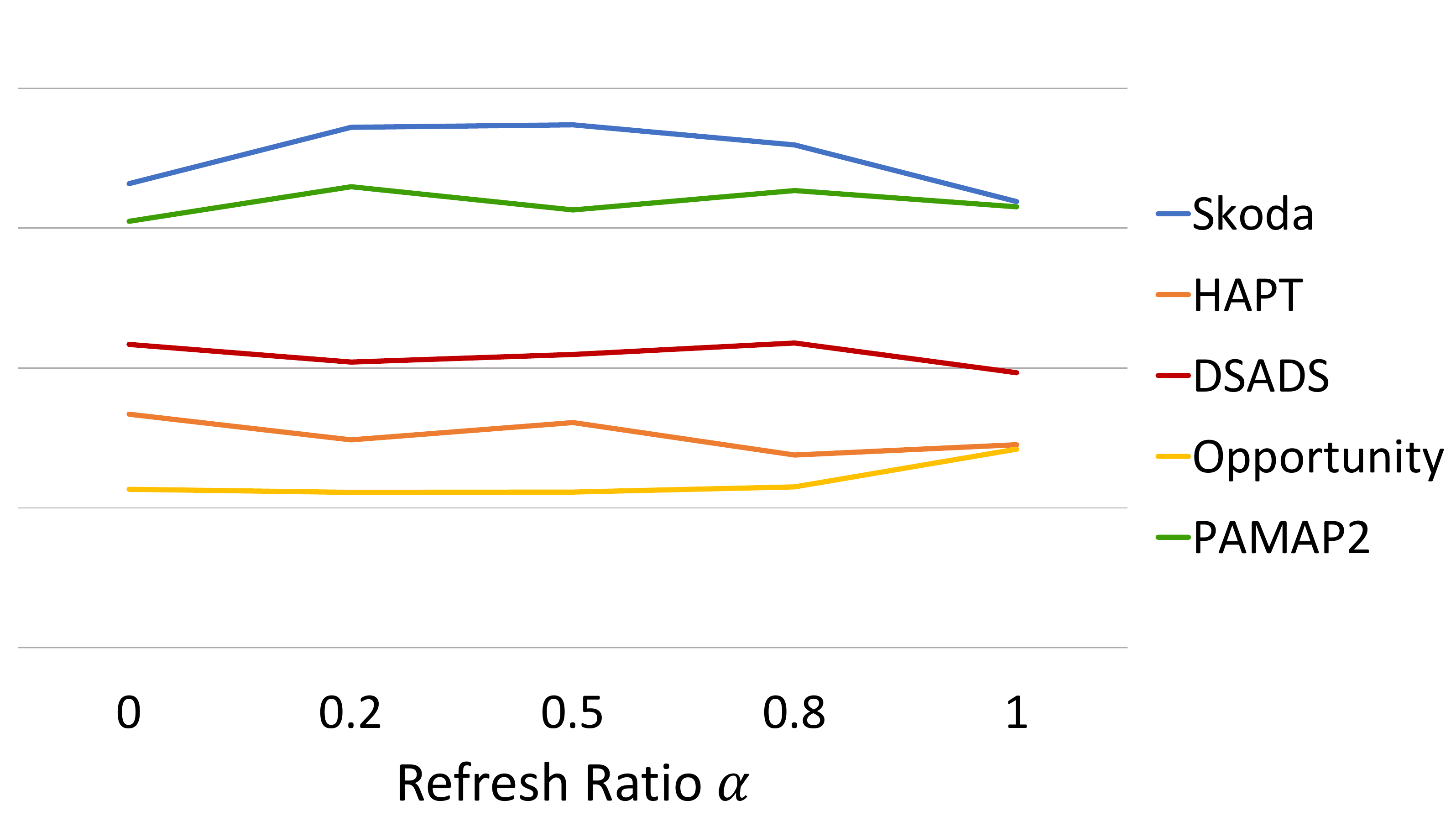}
        \caption{Overall Performance}
    \label{fig:vary_alpha_overall}
    \end{subfigure}
    \caption{Effect of refresh ratio $\alpha$ on (a) base, (b) new, and (c) overall performance across all datasets.}
    \label{fig:vary_alpha}
\end{figure}{}

\subsubsection{Effect of Refresh Ratio}
\label{sec:refresh_ratio}
The refresh ratio, employed in our continual replay-based prototype adaptation, allows a stabilizing update of the prototypes to incorporate the current new latent space while also preserving information from the previous latent space. Analyzing Equation \ref{eq:weighted_update}, a large $\alpha > 0.5$ gives more emphasis to the previous embedding space, while $\alpha < 0.5$ pushes the prototype embeddings to the new latent space. We analyze its effect on the performance measures in Figure \ref{fig:vary_alpha} while keeping all other parameters fixed.

We observe that $\alpha$ does not have a linear relation with the performance measures, and this can be attributed to the differences in data characteristics and how classes and data points interfere with each other during learning. Analyzing the results, we observe that, for $\alpha = 0$, the \textit{Base} performance for HAPT and DSADS is at its highest at around 53\% while for Skoda it is lower compared to $0<\alpha<0.5$ by a margin of $\sim 7\%$. As $\alpha$ increases beyond 0.5, \textit{Base} performance drops reaching its lowest at $\alpha = 1.0$ for Skoda and DSADS and at $\alpha = 0.8$ for HAPT. Opportunity experienced an increase in \textit{Base} performance with $\alpha > 0.5$ reaching its highest at 0.8 while also experiencing a decrease in \textit{New} performance. This follows the intuitive hypothesis that putting more emphasis on previously learnt embedding space reduces catastrophic forgetting of base classes but increases intransigence when learning new classes. However, it is not always straightforward as it is affected by the faithfulness of the new latent space being learned in representing the different class clusters learnt so far. If at a specific time step, the model update caused a drop in its ability to cluster the different classes, adapting the prototypes to this latent space will hinder its recognition ability making subsequent learning of new classes even more challenging. This can be seen in the Opportunity dataset where setting $\alpha=1.0$ maintained a high \textit{Base} performance while also resulting in an increase in \textit{New} performance for Opportunity and ultimately in the \textit{Overall} performance. For PAMAP2, on the other hand, increasing $\alpha > 0.5$ decreased \textit{Base} performance but increased \textit{New} performance reaching its highest at $\alpha=1.0$. This shows that further analysis of the learning behavior of the model at every time step is critical to better understand the changes that happen at the model and prototype level and in turn improve the overall continual learning process.   

\subsection{Effect of Data and Class Sequence}
Continual learning has proven to be a challenging problem where a model has to learn to deal with several challenges such as catastrophic forgetting and concept drift. Thus, researchers tackling this problem often simplified it to an incremental learning of classes divided into a well-defined, equally divided sequence of tasks \cite{delange2021continual}. While it may sound artificial, this allowed researchers to develop methods that tackle one problem at a time as well as enabled having a defined experimental protocol for comparing different approaches. However, the influence of class and data orderings in the evaluation on such incremental learning systems has received very little attention. 

Inter-class similarity is a key challenge in HAR datasets where some activities can have very similar sensor patterns. For example, DSADS includes the activity classes \textit{lying on right} and \textit{lying on back} that can lead to confusion during the model's learning process. Even during offline training, the model wrongly classifies one class with the other. Continual learning is even more challenging since training is done online with only a subset of classes and limited data samples encountered at every time step. Beyond class orderings, data ordering will also have an effect on the learning process depending on whether a gradual or abrupt distribution shift is observed. This is especially true in an online task-free learning setting. Thus, studying the effect of continual learning compared to online learning when a gradual or sudden change in data distribution is observed would give a deeper understanding about the benefits and effectiveness of such systems. 

Prior work on HAR-based continual learning have typically dealt with the randomness of class or task orderings by conducting several experimental runs and averaging the results. Following the same procedure, we observed how performance varied with different class and data orderings. In some cases, performance exhibited a drop which can indicate that more confusion was encountered leading to a more difficult classification task. While analyzing results on multiple class orderings provides a fairer comparison, we aim to investigate, in future work, the impact of class and data orderings in a task-free continual learning setting which can enable better understanding of the continual learning process and the development of more robust systems \cite{masana2020class}.    

\subsection{Comparison to Prior Work}

In this section, we discuss how our approach compares to other methods, more specifically to other CL analysis applied to HAR \cite{JHA2021}. As previously mentioned, task-incremental learning have dominated prior work. Methods developed in the task-based sequential learning setup all often depend on knowing the task boundaries, which give indications of when to distill knowledge after learning a task. Moreover, only data for the current task is observed before moving to another task which guarantees shuffled i.i.d data and convergence to the current task. On the other hand, in a task-free continual learning setting, small non-i.i.d batches are streamed that include data from old as well as new classes. This posed a challenge when comparing our framework to other CL methods in HAR.     

In terms of prototype-based classification, the closest method that uses prototypes in class-incremental continual learning was proposed in iCARL \cite{rebuffi2017icarl}. However, the main difference is that it depends on task information during training. More specifically, iCARL stores new task prototypes or exemplars to represent the class mean in feature space. These, in turn, are used for nearest mean prediction and knowledge distillation. Prediction is only possible after fully learning the task at hand. Thus, task boundaries information is needed to apply knowledge distillation when moving to a new task and for storing class exemplars of previous tasks. Moreover, iCARL requires recalculating all prototypes every time for prediction which is computationally expensive. On the other hand, \textit{LAPNet-HAR} performs online evaluation as well as training using the prototypes in an online task-free way. The prototypes are continually refreshed and updated over time eliminating the need to recalculate them every time. 

Another branch of work focused on model expansion in a task-free setting, such as CURL \cite{rao2019continual} and CN-DPM \cite{lee2019neural}, while not explored and benchmarked in HAR. CURL enables unsupervised task-free representation learning using Variational Auto-Encoder with generative replay. CN-DPM follows a Dirichlet process to train a set of experts for subsets of the data. Its key contribution is the unbound dynamic expansion of the learner. \textit{LAPNet-HAR} evades unbound allocation of resources by keeping the network capacity static but instead incorporates prototypes for new classes which are d-dimensional and fixed in size.     

\section{Lifelong Learning Application Scenarios}

Lifelong learning opens up new possibilities in a wide range of applications. In real-world scenarios, systems built with adaptive models can successfully respond to dynamic changes in data and environments, greatly improving their resilience and resulting in rich user experiences. In this section, we provide examples that demonstrate the benefits of continuously adaptive models in human-activity recognition applications. 

\subsection{A Lifelong Exercise Coach}

Wearables that can track physiological measures have become popular in the last few years. Some of these devices, e.g., smartwatches, are now capable of passively recognizing and quantifying a limited set of physical activities such as walking, running, swimming and biking. Consider a future version of these devices that serves as a lifelong exercise guide spanning several decades. The devices might employ pre-trained models for common activities but it is crucial that they learn to recognize new forms of exercises and adapt. A runner might start with short races at first but become more sophisticated over the years and add strength training to her weekly regimen in order to gain speed, avoid injuries, and tackle longer races. Years into running, the same runner might decide to enroll in a yoga class for recovery after long-runs while also adding a weekly session of plyometrics to gain lower-body power. Later in life, due to a career change, move to a different country, or health issue, the individual might abandon running in favor of a completely different sport such as badminton or golf. Ideally, all of these activities should be monitored by a continuously adaptive model powering a recommendation engine that dispenses feedback and guidance about training, form, nutrition, sleep and other relevant lifestyle factors.

\subsection{Long-Term Health Monitoring for Independent Living}

A long-term goal of the field of ubiquitous computing has been to develop new sensing, communication, and modeling technologies that enable older adults to live independently in their homes. This vision often requires systems that can track and assess activities of daily living (ADL) in real-world environments and for long periods of time. While many approaches have been successful in recognizing human activities from samples of activity classes, they operate under the prevailing assumption that human behaviors, activity routines, and environmental settings are static and do not change over time. Needless to say, this assumption is misaligned with the dynamics of real life. As individuals grow older, they often develop disabilities that cause them to perform activities differently (e.g., change in walking gait). New hobbies (e.g., gardening) and wellness activities (e.g., resistance exercises to combat osteoporosis), might become part of the everyday routine at the same time that mobility patterns change (e.g., less driving). In light of these anticipated changes, having to retrain an activity recognition and monitoring model from scratch on new collected data is not desirable, if not altogether practical.

\subsection{Smart Home and Accessibility}

Advances in machine perception combined with lifelong learning have the potential to enable new applications in accessibility. Imagine a smart home that can aid mobility-impaired individuals by identifying cues that they are about to perform primary tasks (e.g., take a shower) and assisting them with secondary tasks (e.g., automatically turn on bathroom lights, adjust water temperature, close bedroom curtains). Similarly, consider a system for blind individuals that provides discreet auditory feedback about what others are doing around them, giving them improved awareness of their surroundings. It is common for visual, hearing, mobility and cognitive impairments to worsen over time. Therefore, offering appropriate and adaptive levels of support and scaffolding are key to realizing these scenarios in a sensible and successful manner. Adaptive and lifelong modeling is key. An additional challenge in the context of smart homes is that physical environments differ significantly, whether due to layout differences or background noises (e.g., ambient hum of air conditioners, urban sounds of city living). Sometimes, environmental characteristics might change unexpectedly and non-deterministically, such as due to construction or a move. Thus, having a mechanism to continuously adapt and personalize to the target environment over the long term is a crucial component to provide a smooth and continuous experience.

\section{Limitations and Future Work}

In this paper, we presented \textit{LAPNet-HAR}, a lifelong adaptive learning system that is capable of continually learning from streaming data in a task-free setting. We regard our framework as the first step towards truly continual learning in HAR in naturalistic settings. However, lifelong learning has a wider scope and includes several challenges that remain untackled. In this section, we discuss several ways we plan to extend our framework to ultimately develop an end-to-end continual learning system for HAR.

As a first step in investigating the effectiveness of our \textit{LAPNet-HAR}, we made assumptions regarding the ability to discover new activities from streaming data as well as the availability of high-quality ground truth activity labels. When new activities emerge over time, an obvious challenge is being able to determine when a new activity is encountered. Leveraging our framework's clustering approach with class prototypes, a threshold-based activity discovery mechanism over the metric space \cite{ren18fewshotssl} can be incorporated. This can further prevent corrupting previously learned prototypes. Moreover, with forming new class prototypes, analyzing measures such as cluster's density, weight, and temporal characteristics can help eliminate unwanted outliers. Assuming activity discovery is enabled, another challenge is acquiring the ground truth labels. In reality, acquiring annotations has been well acknowledged as time- and effort-consuming with several work being devoted to acquiring data labels while reducing user burden \cite{intille2016muema,9156226}. Furthermore, acquiring annotations is not only about \textit{how} to annotate but also about \textit{what} data to annotate and \textit{when}. All three questions ultimately alleviate the labelling effort. Active learning directly aims to answer these questions by selecting the most informative data samples to annotate \cite{10.1145/3351228}. Adaimi and Thomaz \cite{10.1145/3351228} demonstrated how querying and annotating the most uncertain data samples to use for training reduces the labeling effort while maintaining model accuracy. 

As the performance for continual learning on HAR still has room for improvement, in the future, we will look into techniques to further mitigate catastrophic forgetting and also reduce the interference between different classes. To do so, we will look into another important question that deals with the idea of whether and when model adaptation is really needed. Previous work have more often assumed that, at every time step, the model is updated with a new batch of data. However, in reality, not all new data encountered is adding valuable information, and thus can effectively be ignored. This would further reduce catastrophic forgetting since with every incremental learning step, the model is more prone to forgetting due to the increasing number of intermediate model updates. Using the prototypical framework proposed, we will analyze when updating the model weights is needed. This can be done by studying the rate of change in class prototypes as well as monitoring distribution shifts. Last but not least, the prototype memory stores prototypes for every class seen so far. However, in some cases, an activity class might not be encountered again. For example, a student stops riding a bike after buying a car, or an athlete stops playing football after an injury. Thus, it makes sense for the model to forget an old activity that will never be encountered again and make room for a new one. Therefore, we plan to investigate an aging process where prototypes of outdated classes are disposed of. 

In this work, we aimed to demonstrate the promising performance of our framework on several widely-used datasets, with activities ranging from 10 up to 19 classes. However, a more longitudinal real-world analysis would provide insights to additional challenges and open research questions, especially in HAR. In some cases, continual learning tends to slow down the adaptation to newly seen data, as opposed to pure online learning, as it tries to preserve previously learnt knowledge and stabilize the learning process. More specifically, when the new data is much more informative or representative than the old, continual learning can have a negative effect on the training. For example, imagine a user went on vacation to a new environment. Pure online learning would adapt fast to this new environment as opposed to continual learning that would exhibit a slower adaptation. Essentially, the benefit of the stabilizing effect of continual learning depends on the time scale of the changes in the data. Thus, a large-scale longitudinal analysis would push this field further towards an end-to-end sustainable real-world systems. In the future, we aim to collect a longitudinal dataset, from which we can define a data streaming protocol more representative of the real-world that researchers can refer to. Ultimately, this would enable researchers to extend the current continual learning field for HAR.

\section{Conclusion}

This paper proposes \textit{LAPNet-HAR} to support task-free continual learning from data streams in human activity recognition with prototypical networks. \textit{LAPNet-HAR} enables online learning of unlimited number of activities by relying on prototype-based classification. Catastrophic forgetting and model intransigence are addressed by combining experience replay and dynamic prototype adaptation. Inter-class separation is enforced during continual learning using a margin-based contrastive loss. With extensive experiments on 5 HAR datasets, \textit{LAPNet-HAR} has demonstrated promising results for supporting task-free continual activity recognition. We empirically investigated the efficacy of our prototypical approach on 5 HAR datasets in task-free data incremental scenarios. We hope this work would encourage research in this direction, focusing on applications beyond classification or task-incremental learning. Ultimately, this work is considered as a natural next step in moving beyond offline learning and towards lifelong learning which will enable the large-scale and long-term deployment of mobile recognition systems.